\documentclass{bmvc2k}

\usepackage{tikz}
\usepackage{graphicx}
\usepackage{booktabs}
\usepackage{caption}
\usepackage{hyperref}
\usepackage{amssymb}
\usepackage[capitalise]{cleveref}
\usepackage{url}
\usepackage{enumitem}
\usepackage{xspace}
\newcommand{\methodname}{PoseDreamer\xspace}
 
\title{\methodname: Scalable and Photorealistic Human Data Generation Pipeline with Diffusion Models}

\addauthor{Lorenza Prospero}{lorenza@robots.ox.ac.uk}{1,2}
\addauthor{Orest Kupyn}{okupyn@robots.ox.ac.uk}{2}
\addauthor{Ostap Viniavskyi}{viniavskyi@ucu.edu.ua}{3}
\addauthor{Jo\~{a}o F. Henriques}{joao@robots.ox.ac.uk}{2}
\addauthor{Christian Rupprecht}{chrisr@robots.ox.ac.uk}{2}

\addinstitution{
 The Podium Institute for Sports Medicine and Technology\\
 University of Oxford\\
 Oxford, UK\\
}
\addinstitution{
 Visual Geometry Group\\
 University of Oxford\\
 Oxford, UK\\
}
\addinstitution{Ukrainian Catholic University \\
Lviv, Ukraine}

\runninghead{Prospero, Kupyn, Viniavskyi, Henriques, Rupprecht}{PoseDreamer}

\begin{document}
\maketitle
\begin{figure*}
\includegraphics[width=\textwidth]{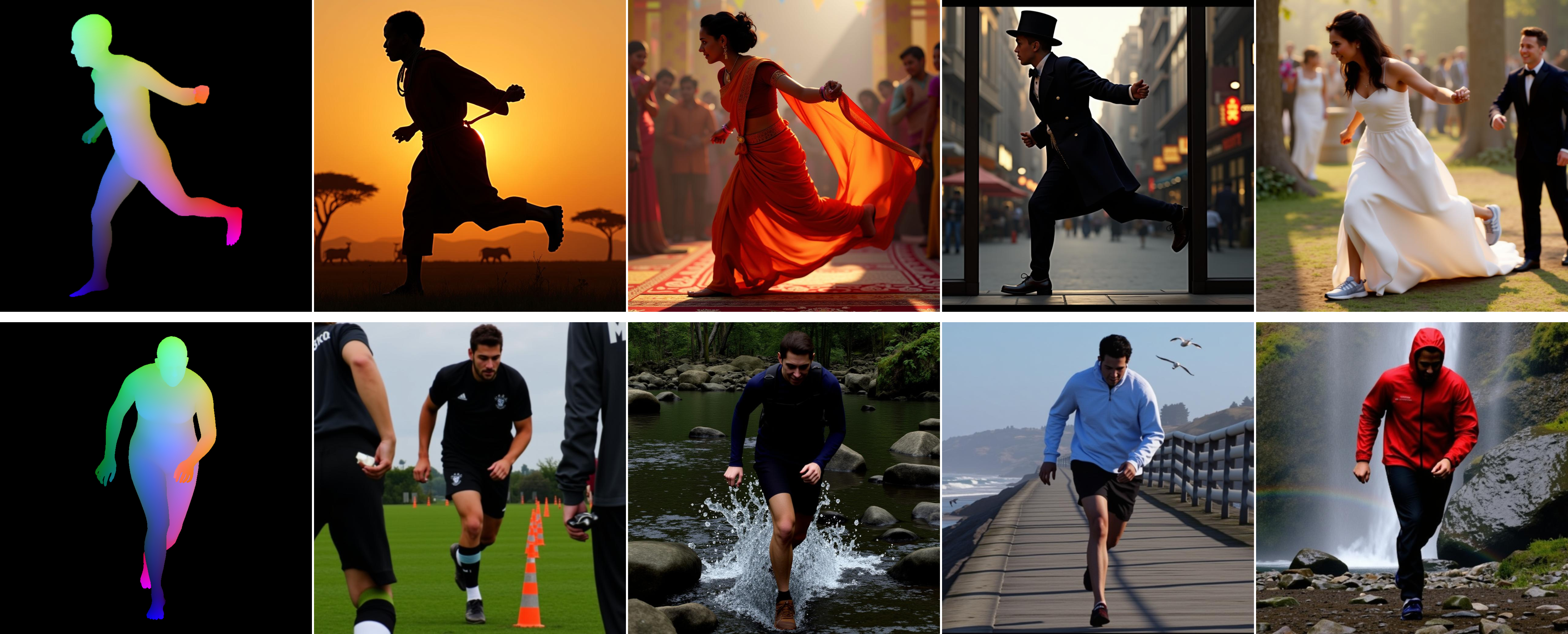}
\centering
\caption{\textbf{\methodname Pipeline:}
Our method builds a high-quality synthetic human dataset using aligned controllable diffusion, hard sample mining, and filtering. It produces photorealistic images with precise spatial control and diverse scenarios for robust training. With fixed SMPL-X~\cite{smplx} parameters, varying only the text caption yields diverse, controllable changes in background, clothing, and environment.}
\label{fig:iamge_diversity}
\end{figure*}
 
\vspace{-0.05cm}
\begin{abstract}
Acquiring labeled datasets for 3D human mesh estimation is challenging due to depth ambiguities and the difficulty of annotating 3D geometry from monocular images. Existing datasets are either real, with manually annotated 3D geometry and limited scale, or synthetic, rendered from 3D engines that provide precise labels but suffer from limited photorealism, low diversity, and high production costs.
In this work, we explore a third path: generated data. We introduce \methodname, a pipeline that leverages diffusion models to generate large-scale synthetic datasets with 3D mesh annotations. Our approach combines controllable image generation with Direct Preference Optimization for control alignment, curriculum-based hard sample mining, and multi-stage quality filtering. Together, these components maintain correspondence between 3D labels and generated images while prioritizing challenging samples to maximize dataset utility.
Using \methodname, we generate more than 500,000 high-quality samples, achieving a 76\% improvement in image-quality metrics over rendering-based datasets. We find that diffusion-generated and rendered synthetic data are complementary rather than competing: generated data, with its photorealism and in-the-wild variability, benefits performance on in-the-wild scenarios, while rendered data retains an advantage on metrics emphasizing 3D-geometric precision. Combining the two yields the best overall results, and we show this complementarity holds across HMR architectures and training objectives.
\end{abstract}
\section{Introduction}
\label{sec:intro}
Human recognition tasks in Computer Vision encompass a range of methods for detecting and describing humans in images. These span simpler tasks, such as bounding-box detection, 2D pose estimation, and semantic segmentation, which identify the position and semantics of humans, to more complex challenges, such as 3D detection, 3D pose estimation, and human avatar reconstruction, which recover the 3D geometry of the human body from images. The success of these methods largely depends on the availability of large, labeled datasets for supervised model training. The Microsoft COCO dataset~\cite{mscoco}, for example, has significantly advanced 2D human pose estimation. However, creating large labeled datasets is costly and time-consuming, and labeling 3D geometry is more challenging still and prone to error. 

A common workaround is to scrape large-scale image collections from the internet and generate pseudo-ground-truth fits. Examples include Instagram images in InstaVariety~\cite{kanazawa2019learning}, and YouTube videos in AVA~\cite{ava}. However, because these images depict real individuals and are not owned by the dataset creators, such datasets raise privacy and consent concerns and can typically be distributed only as links to the original content. This makes them fragile: as hosting platforms remove or relocate content, the dataset degrades over time.

Advanced rendering engines offer one direction, allowing researchers to generate synthetic human datasets with corresponding 3D meshes~\cite{agora, bedlam, psphdri, ultrapose}. However, these approaches face practical limitations that restrict their adoption. First, creating photorealistic renders requires substantial technical expertise in 3D modeling, lighting, and material design. Second, building diverse scenes requires extensive libraries of 3D assets and sophisticated modeling of hair, expressions, and deformations, all computationally intensive. Critically, rendered images often exhibit a ``synthetic look'' that creates a domain gap when models trained on such data are applied to real-world scenarios, limiting their practical use.

In this work, we explore an emerging alternative: generated data. This approach addresses the concerns raised above. Because the images are synthesized rather than depicting real people, they raise no consent issues, and the dataset can be distributed directly rather than as links to third-party platforms. Generated data also avoids the technical barriers of rendering pipelines and, because large generative models are trained on internet-scale real images, it narrows the domain gap that limits rendered data. This raises the question: has the quality of modern generative models crossed the threshold at which their outputs become useful supervision for downstream 3D tasks? Answering it is the focus of this work.  Doing so, however, requires overcoming a fundamental challenge: ensuring precise correspondence between generated images and their 3D annotations, which existing generative models do not inherently satisfy.
\begin{figure}
    \centering
    \includegraphics[width=\textwidth]{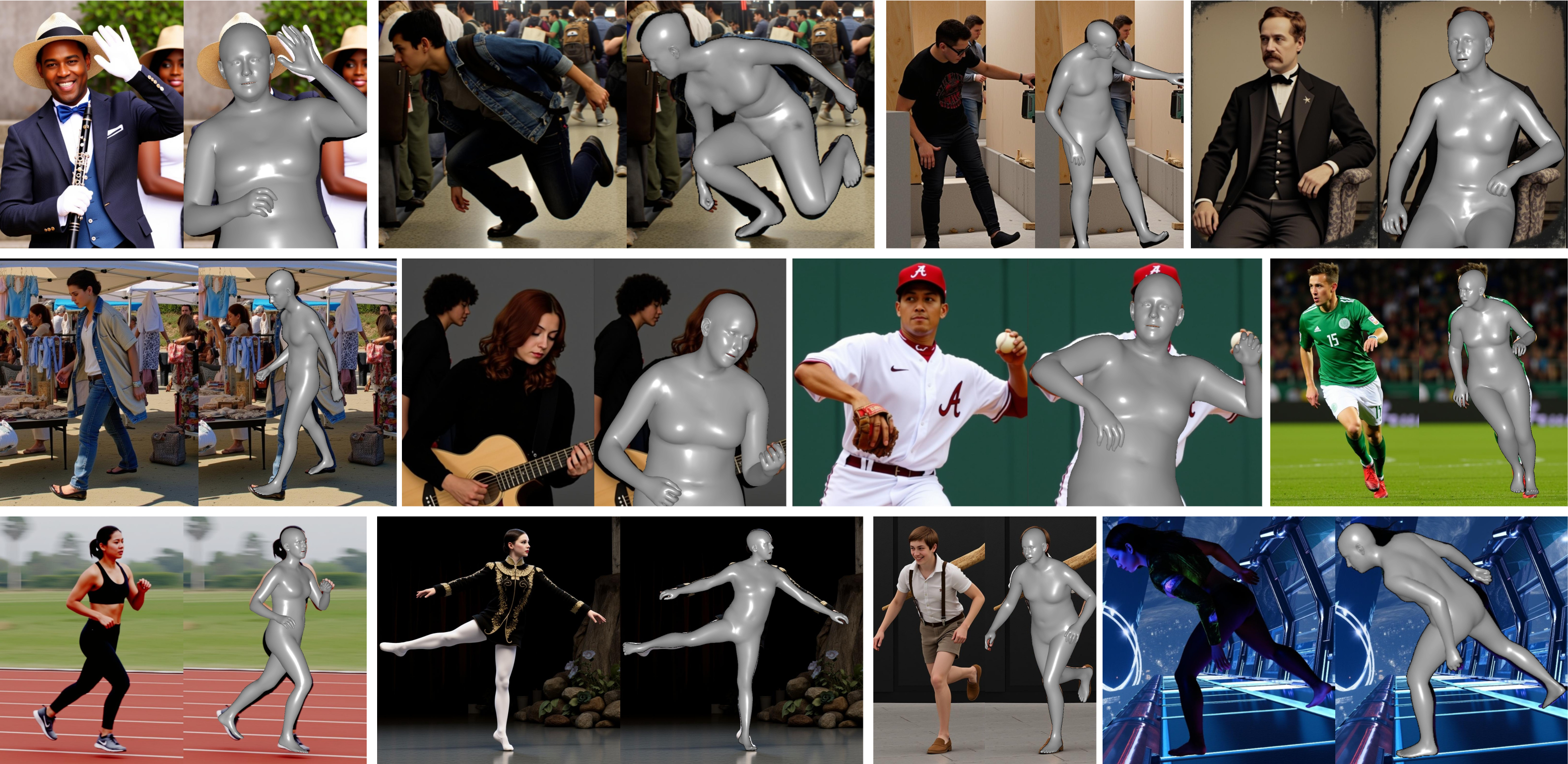}
    \captionsetup{type=figure}
    \caption{\textbf{Examples from the \methodname Dataset.}
High-quality synthetic human samples generated using our pipeline. The examples highlight photorealistic appearance, precise spatial control, and diverse, challenging scenarios that support robust model training.}
    \label{fig:teaser}
\end{figure}
We present \methodname, a pipeline for generating synthetic datasets for human mesh recovery that harnesses state-of-the-art diffusion models~\cite{flux} while ensuring precise 3D-to-2D consistency. We show that diffusion-based generation is a practical alternative to traditional rendering pipelines, offering greater scene diversity and visual realism, and that models trained on \methodname reach performance competitive with those trained on much larger synthetic datasets produced at substantially higher cost. Crucially, we characterize a complementarity between generated and rendered data: generated data improves in-the-wild robustness, while rendered data offers geometric precision; combining them outperforms either alone, a pattern that holds across architectures. Examples of the generated images and annotations are shown in Figure~\ref{fig:teaser}.
The contributions of this work are: 
\begin{itemize}[nosep,leftmargin=*]
\item \textbf{Precise Controllable Human Generation:} We develop an approach for generating photorealistic human images with 3D pose control, introducing an enhanced mesh-to-RGB encoding scheme and employing Direct Preference Optimization to align the control model for improved 3D-to-2D consistency.
\item \textbf{Curriculum-Based Generation with Model Feedback:} We introduce a two-stage pipeline that incorporates feedback from downstream mesh recovery models, prioritizing challenging samples through hard sample mining and avoiding redundant easy cases.
\item \textbf{Large-Scale Synthetic Dataset:} We release a synthetic dataset of over 500,000 images,\footnote{\url{https://prosperolo.github.io/posedreamer}} each annotated with a detailed 3D body mesh, providing a diverse resource for training and evaluating human mesh recovery models.
\item \textbf{Analysis of Generated vs.\ Rendered Data:} We validate this complementarity across multiple benchmarks, architectures, and training objectives, showing that affordable generation can complement and partially replace costly synthetic data acquisition.
\end{itemize}
\section{Related Work}


We divide the literature review into related methods on data generation with diffusion models and an overview of datasets for human mesh recovery.

\textbf{Data Generation with Diffusion Models.} 
Diffusion models \cite{flux, sd3} have transformed data generation methods.
Recently, \cite{imagenet_diversity, fake_imagenet, stablerep, synth_imagenet, imagenedtd} improved the performance of classification models by generating synthetic data with latent diffusion models \cite{ldm}. However, the methods are limited to image classification. Diffumask \cite{diffumask}, DatasetDM \cite{datasetdm} and S3OD \cite{kupyn2026s3od} utilize large diffusion models to simultaneously generate synthetic images and annotations for semantic segmentation or depth prediction tasks. 
Instance Augmentation \cite{instance_aug} generates separate objects in an image, providing a framework to augment and anonymize datasets. VGGHeads \cite{vggheads} introduces a first large-scale dataset for human-centric tasks generated with diffusion models. Yet, the data generation process simply predicts ground-truth annotations using state-of-the-art methods, thereby limiting performance. 
Several recent methods also generate diffusion-based data for human tasks. HumanWild~\cite{diffhuman3drecon} shows that adding generated data (using a SDXL-based ControlNet) to CG-rendered datasets improves performance through greater identity and scene diversity.
Diffusion-HPC~\cite{diffusion-hpc}, built on Stable Diffusion, addresses the anatomical inconsistencies often produced by diffusion models through a pose-aware refinement stage. For domain-specific settings, it additionally uses an image-to-image variant that initializes generation from real target-domain images. STAGE~\cite{stage} builds a similar 3D-pose-conditioned generator but targets the auditing of pose estimators rather than producing training data. Finally, Hayakawa et al.~\cite{diff_egocentric} propose a generated dataset for egocentric 3D human pose estimation. 
In contrast to these approaches, we generate images using mesh-to-RGB conditioning for FLUX, achieve standalone performance competitive with rendered data, and characterize the complementarity between generated and rendered data.

\textbf{Real-World HMR Datasets.} Existing datasets with 3D human annotations face significant limitations despite sophisticated acquisition setups. Human3.6M~\cite{human36m} captures 3.6 million poses using motion capture systems but restricts data to 11 actors performing seventeen scenarios in controlled environments. Similarly, 3DPW~\cite{3dpw} provides in-the-wild videos with 3D annotations from video and IMU data, but it is limited to 60 sequences. MPI-INF-3DHP~\cite{3dhp} offers 1.3 million frames from 14 synchronized cameras in laboratory settings. While some provide body-shape parameters, their controlled acquisition severely restricts scenario diversity. Pseudo-labeled in-the-wild datasets relax these constraints but inherit the privacy and accessibility concerns discussed above.
Several datasets provide full avatar reconstruction of the human body obtained either from complex and expensive capturing systems - CAPE \cite{cape} or EHF \cite{smplx}, and have the same pitfalls as the datasets with 3D pose, or use pseudo ground truth data optimized from visual cues, and often suffer from inaccuracies. An example of the latter is NeuralAnnot \cite{neural_annot, 3Dpseudpgts}, which labels avatars using ground-truth annotations, such as 2D and 3D joints, available in other datasets. 

\textbf{Synthetic HMR Datasets.} Advanced graphics pipelines that render 3D human models are another promising approach for collecting large 3D-annotated datasets. AGORA~\cite{agora}, Surreal~\cite{surreal}, and PeopleSansPeople~\cite{psphdri} provide hundreds of thousands of fully annotated images. Other approaches like Ultrapose~\cite{ultrapose} use commercial software but are not publicly available. BEDLAM~\cite{bedlam} is a large-scale synthetic dataset produced using physically plausible cloth simulations and artist-generated textures, requiring substantial monetary and computational investment. In contrast, SynBody~\cite{yang2023synbody} automates generation with procedural assets, reducing cost at the expense of realism. Concurrently, BEDLAM2~\cite{bedlam2} was released, targeting video-based estimation with only modest gains for single-frame HMR; we compare against the established BEDLAM.
In contrast, our method can scale to arbitrary sample counts and scenarios without manual asset collection.
\begin{figure}[t!]
\includegraphics[width=0.99\textwidth]{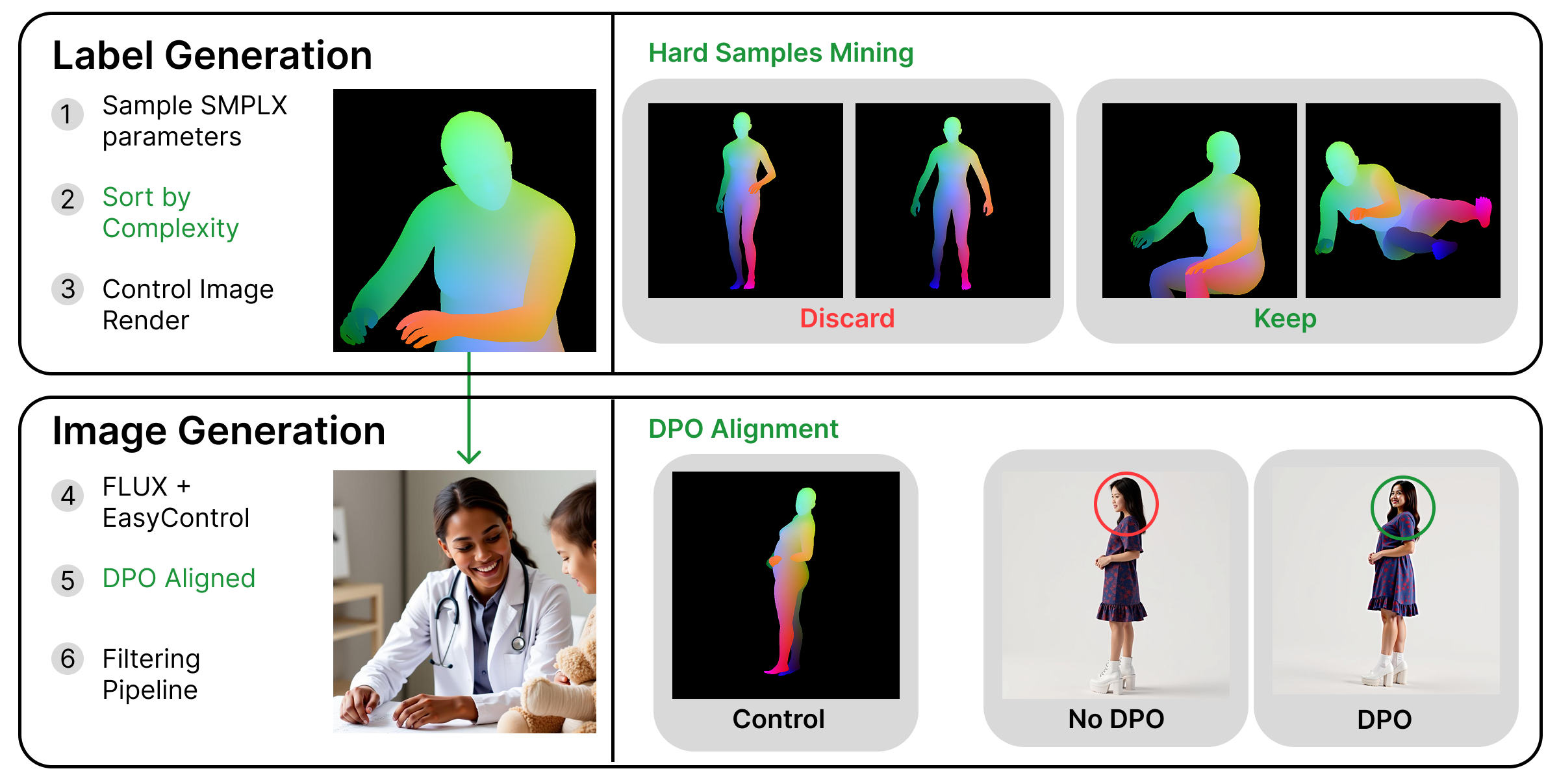}
\centering
\caption{\textbf{Data Generation.} Our pipeline begins with SMPL-X \cite{smplx} parameter and caption generation from multiple sources, followed by curriculum-based hard sample mining to select challenging poses, and concludes with DPO-aligned controllable image generation and comprehensive quality filtering.}
\label{fig:data_pipeline}
\end{figure}

\section{Dataset}
Our objective is to construct a large-scale dataset $\mathcal{D}$ defined as:
$\mathcal{D} = \{(x_i, y_i)\}_{i=1}^{m}$, 
where each pair consists of an RGB image $x_i \in \mathbb{R}^{H \times W \times 3}$ and its corresponding ground truth SMPL-X avatar parameters $y_i$. Following the established paradigm in human mesh estimation~\cite{smpler}, most methods employ a top-down approach: human detection using off-the-shelf detectors, followed by avatar parameter estimation on tightly cropped regions around detected persons. We adopt this framework and focus on single-person mesh recovery.

A straightforward approach would involve generating images $\{x_i\}_{i=1}^{m}$ using latent diffusion models~\cite{ldm}, then applying state-of-the-art mesh recovery models to predict corresponding labels $\{y_i\}_{i=1}^{m}$. However, this strategy has a fundamental limitation: current methods lack 3D-2D consistency. When projected onto the original images, the predicted 3D meshes are frequently misaligned with the actual human poses and body shapes. This prevents the use of predictions as reliable ground-truth labels for downstream model training.

To address this limitation, we propose an inverted generation approach. Instead of generating images first, we begin by sampling diverse SMPL-X~\cite{smplx} avatar parameters $\{y_i\}_{i=1}^{m}$ and utilize these as conditioning signals for image generation. Combined with control-model alignment and extensive filtering, this method ensures consistency between the generated images and their associated 3D body parameters. Examples of diverse scenarios generated using the same control are shown in Figure \ref{fig:iamge_diversity}, while our complete data generation pipeline is illustrated in \Cref{fig:data_pipeline}.

\subsection{Generation Pipeline}

\paragraph{Label Generation.} We generate a set of 3D body model parameters~\cite{smplx} along with detailed captions that describe the corresponding 3D body poses. SMPL-X models human body, hand, and face geometries using a unified parameterization. Specifically, the parameters include pose parameters $\theta \in \mathbb{R}^{55 \times 3}$; body shape parameters $\beta \in \mathbb{R}^{10}$; and facial expression parameters $\psi \in \mathbb{R}^{10}$. The joint regressor $J$ obtains 3D keypoints from parameters via $R_\theta(J(\beta))$, where $R_\theta$ represents the transformation function along the kinematic tree.

A simple approach would randomly sample parameters and condition the generation model on these samples. However, for robust performance and strong generalization, the dataset must contain diverse, realistic scenes with varying interactions, occlusions, and backgrounds, and include a wide range of plausible human 3D poses. To satisfy both requirements, we sample from two complementary sources:

\textit{LAION}~\cite{laion}: This large-scale real-world dataset contains over 5B images with diverse scenes and setups, providing sufficient realism and scene diversity but limited pose variety, with predominantly simple standing, sitting, or walking poses. The pipeline begins with YOLO \cite{yolov8} human detection to identify and crop individual persons from the images. We then employ the Gemma3 vision-language model (VLM) \cite{gemma} to generate detailed captions that describe the visual content, including clothing, environment, and overall scene context. For 3D parameter extraction, we apply a combination of SMPLer-X \cite{smpler} and Token-HMR \cite{tokenhmr} models to predict SMPL-X parameters from the cropped human regions. We deliberately combine predictions from both models to mitigate potential bias that could arise from relying on a single downstream model for 3D mesh parameter prediction. This approach helps ensure more robust and reliable ground truth.

\textit{AMASS}~\cite{amass}: This database contains over 40 hours of diverse human motion-capture data, providing the challenging, dynamic pose diversity that LAION lacks. The dataset includes complex activities such as dancing, martial arts, sports movements, and everyday interactions that exhibit significant joint articulation and body deformation. For each motion sequence, we sample representative frames and use Gemma \cite{gemma} to generate detailed captions with specific body-pose descriptions and comprehensive scene descriptions that incorporate diverse environmental elements, lighting conditions, clothing styles, and background settings. The VLM \cite{gemma} is specifically tasked with describing human pose and generating realistic, detailed scene context, ensuring that the resulting images exhibit wide visual variety while preserving the complex pose structure of the original motion capture data.

\paragraph{Image Generation.} To make ground truth sample parameters compatible with state-of-the-art control models~\cite{easycontrol} and to enforce robust spatial control, we render 3D meshes as RGB images. Common approaches for this conversion include DensePose~\cite{densepose}, which maps body surface coordinates to a 2D texture space, and Continuous Surface Embeddings~\cite{cse}, which encode surface correspondences through learned embeddings that maintain consistency across different poses and body shapes.

However, we found that a simple mapping of vertex IDs to RGB colors yields insufficient visual variation across poses, particularly for distinguishing head orientations and body configurations. This limited color variation prevents generative models from incorporating detailed 3D spatial information during conditioning. Building on prior mesh-to-2D encoding approaches \cite{vitruvian-manifold, ipnet}  and following PNCC notation \cite{pncc} to address this limitation, we propose an improved color coding scheme normalizing each spatial axis (X, Y, Z) independently and mapping the normalized coordinates directly to RGB channels. Compared to naive mapping, it provides richer visual cues that better encode spatial relationships and pose variations.
Our experimental evaluation, reported in the appendix, shows that models trained with this enhanced mesh-to-RGB mapping achieve better 3D pose generation than baseline color-coding approaches, validating the effectiveness of our spatial encoding strategy.

For the control training, we collect a dataset of 130,000 rendered 3D meshes with their corresponding ground-truth images. To ensure accurate alignment between the rendered meshes and images, we leverage annotations from two existing datasets: DensePoseCOCO \cite{densepose} and AGORA \cite{agora}. Using this dataset, we train a spatial control LoRA \cite{lora} following the training procedure of EasyControl \cite{easycontrol}. This approach enables the training of a well-aligned control mechanism for image generation. It also provides the flexibility to combine multiple independently trained LoRAs, allowing for more versatile and compositional control. 

\begin{figure*}[t]
\includegraphics[width=0.99\textwidth]{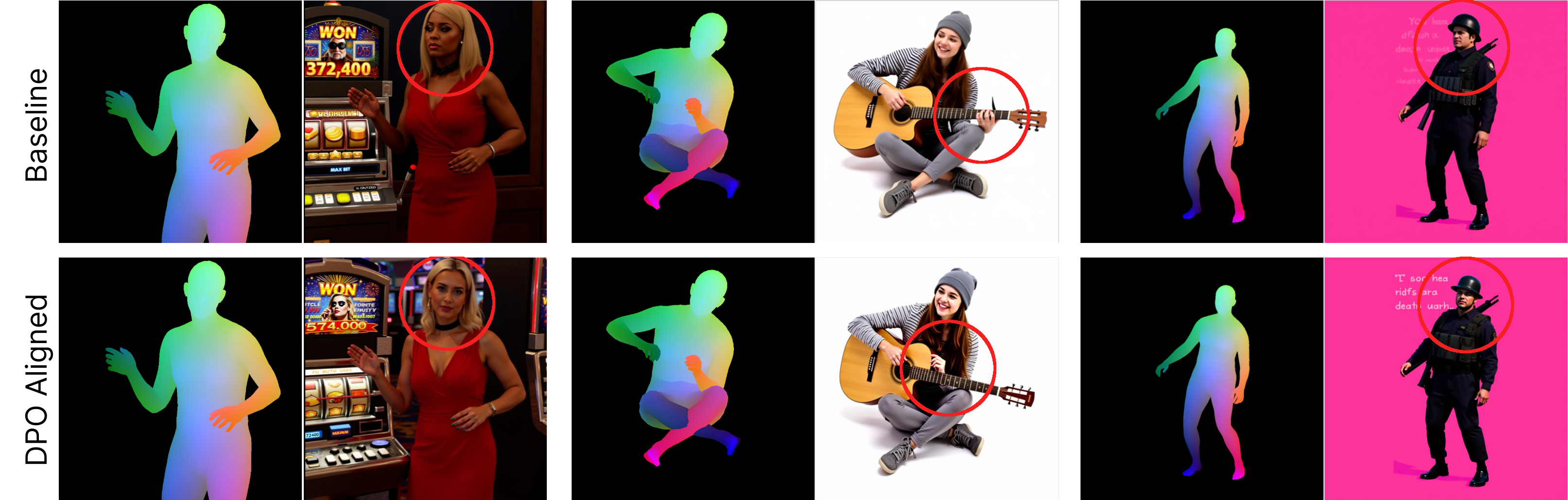}
\centering
\caption{\textbf{DPO Alignment Effectiveness.} Comparison between baseline control model (top row) and DPO-aligned model (bottom row) using identical SMPL-X parameters and captions. The aligned model generates more consistent poses with greater adherence to the ground truth 3D parameters, demonstrating improved control precision and reduced pose deviations.}
\label{fig:dpo_lora}
\end{figure*}

\subsection{Control Model Alignment}
While EasyControl \cite{easycontrol} provides a mechanism for conditioning image generation on 3D mesh inputs, it represents a weak conditioning method that operates through cross-attention mechanisms without explicit spatial constraints. The model learns to associate control signals with visual patterns during training, but lacks architectural guarantees to enforce precise adherence to the provided 3D pose information, often resulting in generated images that deviate from the intended body configurations.

However, we can effectively detect such misalignments using robust 2D proxy metrics. State-of-the-art 2D joint regressors~\cite{yolo-pose} are significantly more robust than their 3D counterparts, as they are trained on larger datasets and tackle a fundamentally easier task of 2D keypoint localization. Our evaluation setup proceeds as follows: given a generated image and the corresponding ground-truth SMPL-X parameters used for conditioning, we first reproject the SMPL-X model's 3D joints to obtain the ground-truth 2D keypoints. We then apply a 2D joint regressor to the generated image to predict 2D keypoints, and compute the Object Keypoint Similarity (OKS) score between the predicted and reprojected keypoints:
\begin{equation}
\text{OKS} = \frac{\sum_i \exp\left(-\frac{d_i^2}{2s^2\kappa_i^2}\right) \delta(v_i > 0)}{\sum_i \delta(v_i > 0)},
\end{equation}
where $d_i$ is the Euclidean distance between predicted and ground truth keypoints, $s$ is the object scale, $\kappa_i$ are per-keypoint constants, $v_i$ indicates keypoint visibility, and $\delta(v_i > 0)$ is 1 for visible keypoints and 0 otherwise. This 2D evaluation approach reliably measures pose correspondence while being significantly more stable than direct 3D mesh regression methods, making it ideal for assessing control fidelity.

This capability to reliably measure control alignment enables us to employ Direct Preference Optimization (DPO) \cite{dpo, diffdpo} to systematically improve the conditioning mechanism. We can enhance control precision and overall generation quality by optimizing the control model to favor generations that exhibit higher pose alignment scores.

We construct preference pairs by ranking generated images according to their OKS scores relative to the ground truth SMPL-X parameters. Images with higher alignment scores are treated as preferred samples ($x_0^w$), while those with lower scores become less preferred samples ($x_0^l$), where $x_0^w \succ x_0^l$.

Following the Flow-DPO framework~\cite{videoreward}, we optimize a control model~with:
\begin{equation}
\begin{aligned}
&-\mathbb{E} \left[ 
\log \operatorname{sigmoid} \left( -\frac{\beta_t}{2} (L_w - L_l) \right) 
\right] \\
&\text{with } 
L_w = \| v^w - v_\theta^w \|^2 - \| v^w - v_{\text{ref}}^w \|^2,
\quad
L_l = \| v^l - v_\theta^l \|^2 - \| v^l - v_{\text{ref}}^l \|^2 \, ,
\end{aligned}
\end{equation}
where $\beta_t = \beta (1 - t^2)$, and $v^*$ denotes the velocity field for rectified flow models. This objective encourages the control model to generate images that more accurately reflect the input 3D pose constraints, thereby improving control precision and overall generation quality.

We train a separate Low-Rank Adaptation (LoRA)~\cite{lora} with rank $r=128$ and $\alpha=128$. This strategy provides the additional advantage of memory efficiency by eliminating the need to store the reference model separately, as the base model with the adapter disabled naturally serves as $p_{\text{ref}}$ during training. Training is conducted with a batch size of 32 for 6,400 iterations using the Adam optimizer with a learning rate of $5 \times 10^{-6}$. The complete fine-tuning process requires 2 days on 4 A6000 GPUs.

We validate the effectiveness of alignment on a 1,000-sample held-out test set by measuring control precision using OKS scores between generated images and ground truth 2D keypoints. The baseline control model achieves an OKS of $0.874 \pm 0.103$, while our DPO-aligned model reaches $0.927 \pm 0.052$. This represents a substantial \textbf{42.1\%} reduction in the error rate, demonstrating significant improvements in pose-alignment consistency and reduced variance in generation quality.  DPO is thus an efficiency mechanism: it increases the pass rate (by 7\% on our held-out subset), avoiding wasted GPU compute on discarded samples. We show examples of the improved OKS alignment in \Cref{fig:dpo_lora}.

\subsection{Hard Sample Mining}
Given an aligned control model and an extensive library of 3D body parameters and captions, we can generate datasets of arbitrary size. However, random sampling often yields redundant samples with similar poses, providing weak training signal. We address this with a curriculum-based two-stage selection that prioritizes challenging examples. 

\textit{Stage 1: Baseline Generation.} We randomly sample $N$ parameter vectors and generate an initial dataset using the aligned control model. We then train an SMPL-X regressor~\cite{smpler} on this dataset and evaluate it on a held-out test set, using OKS between its predictions and ground-truth 2D keypoints. Lower OKS indicates more challenging samples where the regressor underperforms.

\textit{Stage 2: Difficulty-Aware Generation.} We train a gradient boosting decision tree regressor~\cite{lightgbm} to predict OKS scores directly from SMPL-X parameter vectors, letting us rank candidates by expected difficulty without generating images. We then select the top $N$ samples with the lowest predicted OKS and generate the final dataset with the aligned control model. This focuses computation on samples that provide the most learning value while avoiding redundant, easy cases.

This strategy is complementary to pose-source diversity: sources expand the candidate pool, while hard mining selects which candidates to generate given a fixed compute budget.

\subsection{Dataset Filtering}

Despite our high-quality aligned control model, failure cases inevitably arise. To ensure robust data quality, we implement a comprehensive three-stage filtering pipeline that addresses the primary failure modes we observe in generated data: crowded scenes and severe occlusions, general pose misalignment, and 3D head pose inconsistencies.

\textbf{Crowded Scenes.} We run YOLO \cite{yolov8} on generated images and filter out samples containing more than five detected persons. This threshold removes only highly crowded scenes in which pose estimation quality cannot be reliably ensured, thereby allowing the model to learn partial occlusions and human-to-human interactions.

\begin{figure*}[t]
\includegraphics[width=0.99\textwidth]{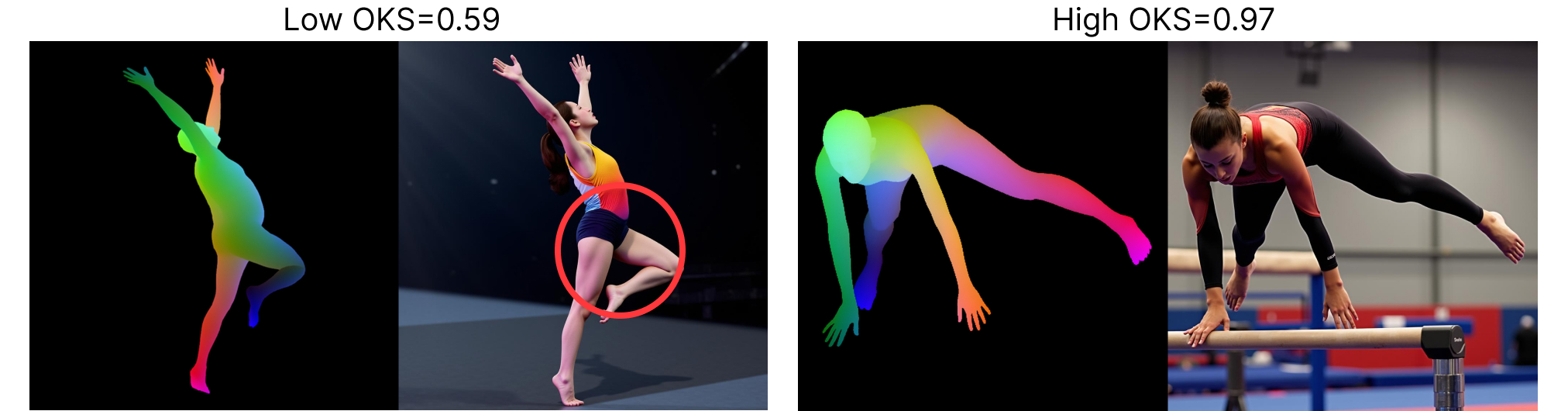}
\centering
\caption{\textbf{OKS scores.} Comparison between a low-OKS example (left) and a high-OKS example (right). Each pair shows the conditioning pose alongside the generated image. In the low-OKS case, the generated image has the left and right legs swapped relative to the conditioning pose, causing the detected keypoints to disagree with the projected mesh joints. Only generations with an OKS score above 0.8 pass our filtering stage.}
\label{fig:oks_examples}
\end{figure*}

\textbf{General Misalignment.} We employ the same OKS-based evaluation framework used in control model alignment to filter out samples with poor pose correspondence. Images with OKS scores below a predefined threshold are discarded, some examples are shown in \Cref{fig:oks_examples}. The combination of alignment and filtering is necessary because control model alignment provides key advantages: (a) computational efficiency - aligned models have significantly higher filtering pass rate, reducing wasted computation on poor samples; (b) quality improvement - samples that pass the filtering threshold exhibit better overall alignment when generated from aligned models; and (c) bias mitigation - unaligned models tend to bias toward simple poses that achieve consistency, potentially excluding challenging poses from the final filtered dataset that are essential for robust training.

\textbf{3D Head Pose Inconsistency.} We extract 3D head pose parameters (roll, pitch, yaw) from the ground truth SMPL-X parameters and use VGGHeads~\cite{vggheads} to predict head pose from generated images. Samples with mean absolute error (MAE) above a threshold in any rotation axis are filtered out, ensuring head pose consistency in challenging out-of-distribution poses where general body alignment metrics may be insufficient.

This multi-stage filtering approach ensures that only high-quality, consistent samples contribute to the final training dataset while maintaining diversity in challenging scenarios. The final dataset consists of 500,000 samples filtered from 800,000 initially generated images (500,000 conditioned on LAION poses and 300,000 on AMASS poses). The full generation process required 6,700 GPU hours. This translates to a modest cost compared to traditional rendering pipelines. At July 2026 cloud rates (\$0.57/hr A6000, \$2.90/hr H100)\footnote{GPU pricing from \url{https://gpuperhour.com}, accessed July 2026.}, the full \methodname\ pipeline (generation + DPO + training) costs $\sim$\$4K. BEDLAM required 111 commissioned simulation-ready garments, cloth simulation, licensed assets, and commercial textures~\cite{bedlam}. Public 3D-artist rates (\$105--250/garment)\footnote{Fiverr 3D artist pricing guide: \url{https://www.fiverr.com/resources/guides/costs/3d-artist}, accessed July 2026.} already imply \$11--27K for garments alone, excluding simulation and asset preparation. Unlike rendering pipelines, \methodname\ scales primarily with compute rather than specialist labor.

\section{Experimental Evaluation}
We evaluate \methodname\ along four dimensions. First, we measure downstream performance by training human mesh recovery models on our generated data and benchmarking their generalization to real-world test sets, both alone and in combination with established synthetic datasets. Second, we compare against existing diffusion-based generation pipelines and show that \methodname\ can be
targeted at specific domains. Third, we assess image quality by comparing \methodname\ against rendered synthetic datasets on standard photorealism metrics. Finally, we ablate the components of our pipeline to isolate the contribution of each design choice. 

\subsection{Human Mesh Recovery Benchmarking}
We evaluate human mesh recovery along two axes: (i) general benchmarking using the SMPLer-X~\cite{smpler} protocol across diverse academic datasets, and (ii) cross-architecture validation under a different HMR model with a different training objective.

\textbf{SMPLer-X results.} We follow the SMPLer-X~\cite{smpler} benchmarking protocol, training the SMPLer-X-S model on our synthetic data and evaluating its generalization across multiple real-world benchmarks with ground-truth SMPL-X annotations, comparing models trained on single datasets and mixtures. We evaluate on five academic datasets: \textbf{AGORA}~\cite{agora}, a synthetic benchmark for rendering engine evaluation; \textbf{UBody}~\cite{ubody}, covering 15 real-world scenarios with diverse in-the-wild captures; \textbf{EgoBody}~\cite{egobody}, a large-scale first-person view dataset; \textbf{3DPW}~\cite{3dpw}, providing challenging outdoor sequences; and \textbf{EHF}~\cite{smplx}, a laboratory dataset with 100 diverse poses.

\begin{table}[t]
  \caption{\textbf{Comparison vs EHPS Datasets.} Top-1 values are bolded,
and the rest of Top-2 are underlined. \#Inst.: number of instances used in training. ITW: in-the-wild. For each dataset, the model is trained on its training set and evaluated on the \emph{val} set of AGORA and \emph{testing} sets of UBody, EgoBody (EgoSet), 3DPW, and EHF. The upper part of the table shows comparison for models trained on multiple datasets, while the lower part of the table shows comparisons for models trained on a single dataset. Datasets used to train the SMPLer-X-S5 model are marked with $\clubsuit$.}
  \label{tab:single_datasets}
  \centering
  \resizebox{\textwidth}{!}{
  \begin{tabular}{lcccccccc}
    \toprule
    & & & 
    \multicolumn{1}{c}{AGORA} &
    \multicolumn{1}{c}{UBody} &
    \multicolumn{1}{c}{EgoBody} &
    \multicolumn{1}{c}{3DPW} &
    \multicolumn{1}{c}{EHF} & \\
    
    \cmidrule(lr){4-4}
    \cmidrule(lr){5-5}
    \cmidrule(lr){6-6}
    \cmidrule(lr){7-7}
    \cmidrule(lr){8-8}
    
    Dataset & \#Inst. & Type &
    PVE$\downarrow$ &
    PVE$\downarrow$ &
    PVE$\downarrow$ &
    MPJPE$\downarrow$ &
    PVE$\downarrow$ &
    MPE$\downarrow$
    \\
    \midrule
    \methodname + BEDLAM  & 750K & ITW Mix & 156.8 & \textbf{97.6} & \textbf{106.1} & \textbf{96.8} & \underline{84.3} & \textbf{108.3} \\
    MSCOCO + BEDLAM & 750K & ITW Mix & \underline{156.1} & \underline{101.5} & \underline{113.4} & \underline{100.6} & \textbf{78.7} & \underline{110.1} \\
    5 datasets ($\clubsuit$) & 750K & ITW Mix & \textbf{119.0} & 110.1 & 114.2 & 110.2 & 100.5 & 110.8 \\
    
    \midrule \midrule
    \methodname & 500.0K & ITW Gen & 176.0 & \textbf{98.4} & \underline{120.3} & 102.7 & \underline{95.2} & \textbf{118.5} \\ 
    BEDLAM~\cite{bedlam} (our rerun) ($\clubsuit$) & 951.1K & ITW Syn & 160.5 & 146.3 & \textbf{109.5} & \textbf{98.7} & \textbf{91.2} & \underline{121.2} \\
    SynBody~\cite{yang2023synbody} & 633.5K & ITW Syn & 166.7 & 144.6 & 136.6 & 106.5 & 112.9 & 133.5 \\
    InstaVariety~\cite{kanazawa2019learning} & 2184.8K & ITW Real & 195.0 & 125.4 & 140.1 & 100.6 & 110.8 & 134.3 \\
    GTA-Human II~\cite{cai2021playing} ($\clubsuit$) & 1802.2K & ITW Syn & \underline{161.9} & 143.7 & 139.2 & 103.4 & 126.0 & 134.8 \\
    MSCOCO~\cite{mscoco} ($\clubsuit$) & 149.8K & ITW Real & 191.6 & \underline{107.2} & 139.0 & 121.2 & 116.3 & 135.0 \\
    EgoBody-MVSet~\cite{egobody} & 845.9K & Indoor Real & 190.9 & 191.4 & \textcolor{gray}{(127.0)} & \underline{99.2} & 101.8 & 142.1 \\
    AGORA~\cite{agora} ($\clubsuit$) & 106.7K & ITW Syn & \textcolor{gray}{(124.8)} & 128.4 & 138.4 & 131.1 & 164.6 & 145.4 \\
    Egobody-EgoSet~\cite{egobody} & 90.1K & Indoor Real & 207.1 & 126.8 & \textcolor{gray}{(103.1)} & 134.4 & 121.4 & 147.5 \\
    RICH~\cite{huang2022capturing} & 243.4K & ITW Real & 195.6 & 168.1 & 137.9 & 115.5 & 127.5 & 148.9 \\
    MPII~\cite{andriluka14cvpr} & 28.9K & ITW Real & 202.1 & 123.9 & 155.5 & 131.9 & 140.8 & 150.8 \\
    MuCo-3DHP~\cite{Mehta2018SingleShotM3} & 465.3K & ITW Real & 187.7 & 185.4 & 146.4 & 119.4 & 134.7 & 154.7 \\
    PROX~\cite{hassan2019resolving} & 88.5K & Indoor Real & 204.1 & 180.3 & 151.8 & 132.5 & 122.5 & 158.2 \\
    UBody~\cite{ubody} & 683.3K & ITW Real & 207.0 & \textcolor{gray}{(78.7)} & 145.6 & 149.4 & 132.1 & 158.5 \\
    SPEC~\cite{kocabas2021spec} ($\clubsuit$) & 72.0K & ITW Syn & \textbf{161.5} & 146.1 & 154.8 & 139.7 & 197.8 & 160.0 \\
    \bottomrule
  \end{tabular}}
\end{table}

\Cref{tab:single_datasets} presents results across these benchmarks. Our diffusion-generated dataset achieves strong performance across all evaluations and yields results comparable to BEDLAM~\cite{bedlam}, which contains roughly twice as many instances and requires substantially greater computational and artistic resources to produce. Notably, our dataset outperforms all real-world datasets in the lower part of the table, which we attribute to the combination of scale, scene diversity, and labels generated natively in the SMPL-X format, whereas real-world datasets typically rely on noisier pseudo-ground-truth fits derived from 2D cues. When combined with BEDLAM, our dataset outperforms all multi-dataset mixtures, including BEDLAM+COCO and the five-dataset ensemble, reflecting complementary strengths: rendered data contributes precise 3D supervision, while generated data contributes photorealistic diversity that rendering pipelines struggle to match. We additionally validate scaling by repeating the evaluation with the larger SMPLer-X-L (ViT-L) backbone, where the same trends hold; full results are provided in the appendix.

\textbf{Cross-architecture validation.} To verify that these findings are not specific to SMPLer-X, we repeat the comparison using HMR2.0~\cite{hmr2}, which employs a ViT-H backbone and a keypoint-dominated training objective. We follow the published HMR2.0 architecture and loss configuration, training each model for 250{,}000 iterations under matched conditions. Each model is trained on the HMR2.0a real-world configuration combined with BEDLAM, \methodname, or both. We use HMR2.0a rather than the full HMR2.0b mix because the additional datasets (InstaVariety, AVA, AI Challenger) suffer from the privacy and accessibility limitations discussed in \cref{sec:intro}. As shown in \cref{tab:hmr2_crossarch}, the same complementarity pattern emerges: BEDLAM leads on 3D-geometric and strict-tolerance metrics (3DPW PA-MPJPE and MPJPE), while \methodname\ and the combined dataset lead on 2D and loose-tolerance metrics (COCO~\cite{mscoco}, LSPet~\cite{Johnson2011LearningEH},
PoseTrack~\cite{andriluka2018posetrack} PCK). The per-metric gaps are smaller than under SMPLer-X, likely reflecting differences in HMR2.0's training configuration. Nonetheless, the appearance-versus-geometry trade-off persists across architectures, supporting it as a property of the data sources themselves rather than a single model artifact.
\begin{table*}[t]
\centering
\caption{\textbf{Cross-Architecture Validation (HMR2.0, ViT-H).} The complementarity pattern holds under a different architecture and training objective: rendered data (BEDLAM) leads on 3D-geometric metrics, while generated data (\methodname) and the combination lead on in-the-wild 2D metrics. Best per column in bold.}
\label{tab:hmr2_crossarch}
\renewcommand{\arraystretch}{1.1}
\resizebox{\textwidth}{!}{%
\begin{tabular}{lcccccccc}
\toprule
 & \textbf{COCO} & \textbf{COCO} & \textbf{LSPet} & \textbf{LSPet} & \textbf{PoseTrack} & \textbf{PoseTrack} & \textbf{3DPW} & \textbf{3DPW} \\
\textbf{Training Data} & \textbf{@.05} & \textbf{@.10} & \textbf{@.05} & \textbf{@.10} & \textbf{@.05} & \textbf{@.10} & \textbf{PA-MPJPE} & \textbf{MPJPE} \\
 & $\uparrow$ & $\uparrow$ & $\uparrow$ & $\uparrow$ & $\uparrow$ & $\uparrow$ & $\downarrow$ & $\downarrow$ \\
\midrule
Real + BEDLAM & 81.82 & 95.18 & 46.33 & 78.11 & 87.00 & 97.02 & \textbf{45.59} & \textbf{72.91} \\
Real + \methodname & 82.44 & 95.27 & 45.73 & 78.27 & 87.94 & \textbf{97.53} & 47.81 & 75.46 \\
Real + BEDLAM + \methodname & \textbf{83.35} & \textbf{95.45} & \textbf{46.49} & \textbf{78.98} & \textbf{88.49} & 97.44 & 46.90 & 74.98 \\
\bottomrule
\end{tabular}%
}
\end{table*}

\subsection{Comparison with Diffusion-Based Baselines}
\label{sec:humanwild_comparison}
HumanWild~\cite{diffhuman3drecon} is, to the best of our knowledge, the only other diffusion-based generic HMR data generation pipeline. It conditions an SDXL-based ControlNet on rendered signals (keypoints, depth, and surface normals) to synthesize images with 3D annotations. As its generated dataset is not publicly released, we perform a matched-scale head-to-head comparison using their publicly released ControlNet: we generate 10{,}000 samples with each pipeline and finetune SMPLer-X-S (pretrained on BEDLAM) on each.
\cref{tab:humanwild} shows that \methodname\ outperforms HumanWild on all three in-the-wild benchmarks and on both image-quality metrics. The largest gap appears on MSCOCO, consistent with the appearance-diversity advantage of \methodname's FLUX-based generation over HumanWild's SDXL base with rendered-normal conditioning. Beyond pose accuracy, the image-quality gap confirms that \methodname\ produces images closer to real-world distributions than the closest publicly reproducible diffusion baseline.

\begin{table}[t]
\centering
\begin{minipage}{0.39\linewidth}
    \centering
  \caption{PVE ($\downarrow$) results for models trained on \methodname and BEDLAM on \textbf{in-the-wild benchmarks}. }
  \label{tab:itw_benchmarks}
  \centering
  \resizebox{\linewidth}{!}{
    \begin{tabular}{l c c c}
      \toprule
      Dataset & UBody & MPII & MSCOCO \\
      \midrule
      BEDLAM \cite{bedlam}
      & 146.3 & 141.1 & 163.7 \\
      \methodname
      & \textbf{97.6} & \textbf{122.3} & \textbf{129.4} \\
      
      \bottomrule
    \end{tabular}%
  }
\end{minipage}
\hfill
\begin{minipage}{0.58\linewidth}
    \centering
    \caption{\textbf{\methodname\ vs.\ HumanWild} at matched 10K scale. SMPLer-X-S (BEDLAM-pretrained) finetuned on each. IS/FID computed on the 10K subsets.}
\label{tab:humanwild}
\footnotesize
\resizebox{\linewidth}{!}{
\begin{tabular}{lccccc}
\toprule
Dataset & MPII $\downarrow$ & MSCOCO $\downarrow$ & UBody $\downarrow$ & IS $\uparrow$ & FID $\downarrow$ \\
\midrule
HumanWild \cite{diffhuman3drecon} & 126.21 & 141.89 & 110.86 & 8.10 & 3.29 \\
\methodname & \textbf{124.20} & \textbf{134.39} & \textbf{101.25} & \textbf{11.84} & \textbf{1.64} \\

\bottomrule
\end{tabular}%
}
\end{minipage}
\label{tab:humanwild_itw_combined}
\end{table}

\subsection{Domain-specific Generation}
\begin{figure*}[h]
\includegraphics[width=0.99\textwidth]{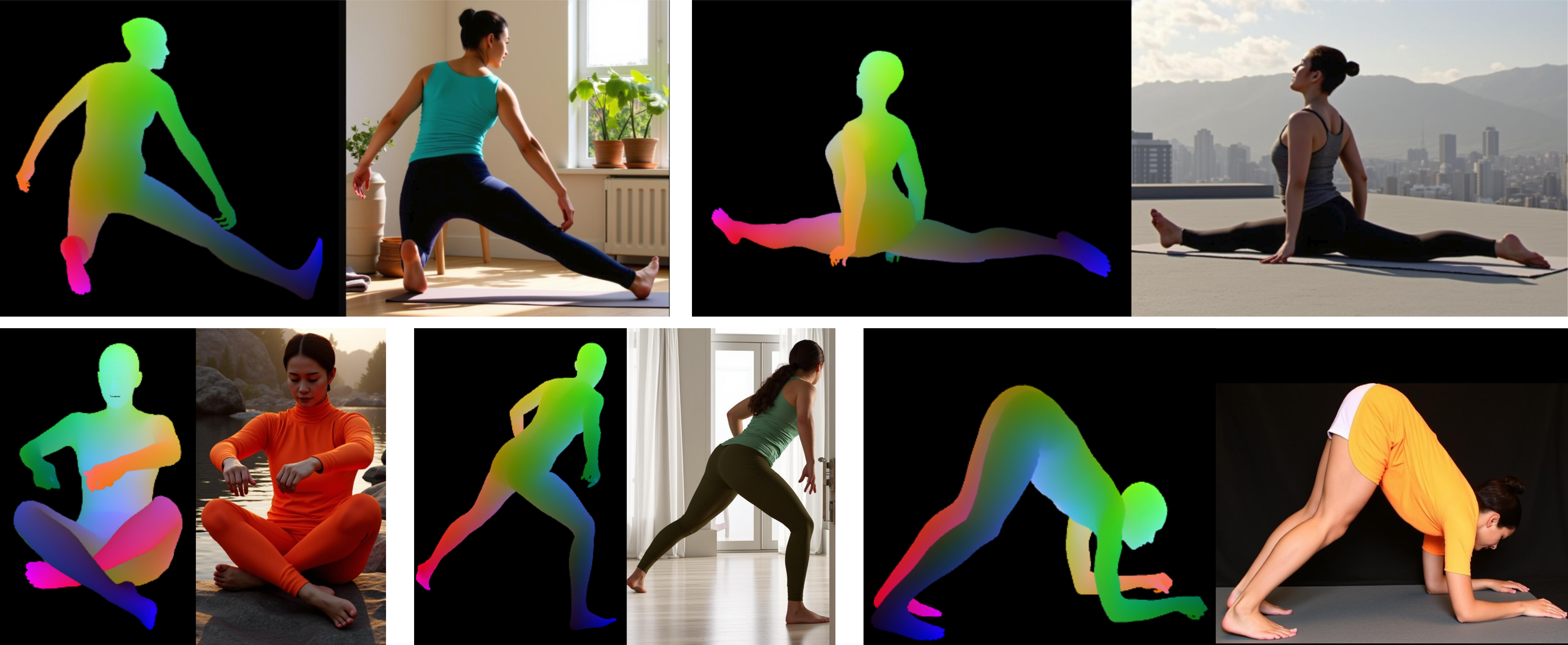}
\centering
\caption{\textbf{Yoga Generations.} Even for the challenging yoga poses from the MOYO~\cite{moyo} dataset, our pipeline generates realistic images with diverse backgrounds.}
\label{fig:yoga-gen}
\vspace{-0.5cm}
\end{figure*}

\begin{figure*}[h]
\includegraphics[width=0.99\textwidth]{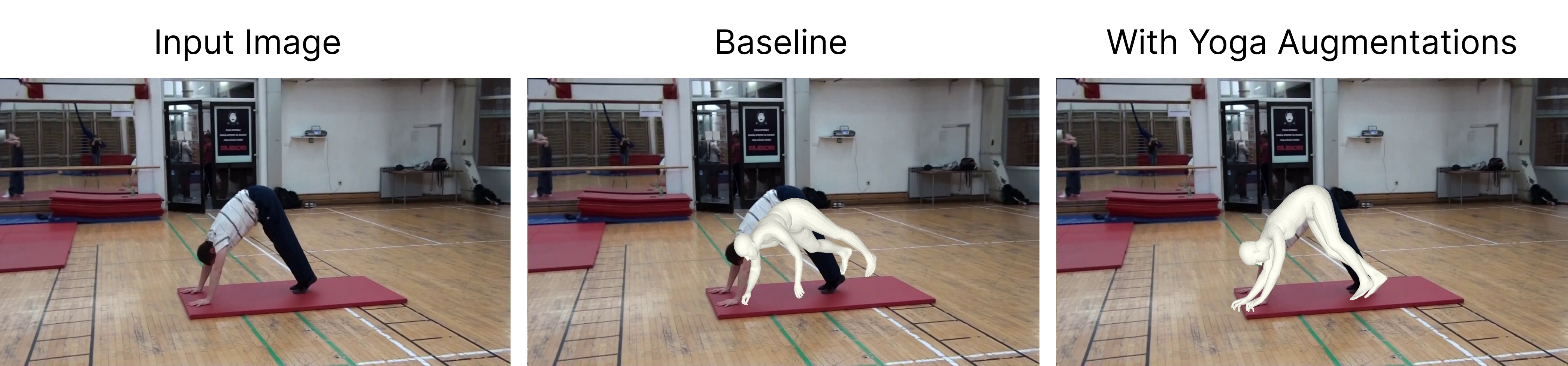}
\centering
\caption{\textbf{Yoga-Augmented Training.} Yoga-augmented training improves pose recovery on MPII yoga images.}
\label{fig:yoga-pred}
\end{figure*}

A key advantage of \methodname\ over traditional datasets is the ability to generate images tailored to a target domain. To evaluate this, we generate 30K images using poses from the MOYO dataset~\cite{moyo} (used exclusively as a pose source, not for retraining), which contains challenging yoga poses underrepresented in other datasets (Figure~\ref{fig:yoga-gen}). We train two SMPLer-X-S models: one on 300K randomly sampled AMASS and LAION poses, and one with 30K of these samples replaced with our yoga-pose images. Both are evaluated on UBody, 3DPW, EHF, and the 487 MPII instances tagged ``yoga'' (Table~\ref{tab:yoga-poses}).
The yoga-augmented model matches the baseline on general benchmarks while decisively outperforming it on the MPII yoga subset, supported by visual comparisons in Figure~\ref{fig:yoga-pred}. Thus, our pipeline generates data tailored to specific domains, improving downstream performance without sacrificing general capability. We additionally compare against Diffusion-HPC~\cite{diffusion-hpc} on the SMART benchmark~\cite{smart} of athletic poses; full results are in the appendix.



\begin{table*}[t]
\centering
\begin{minipage}{0.38\linewidth}
    \centering
    \captionof{table}{\textbf{Ablations.} We evaluate the impact of our pipeline components based on their effect on downstream model performance.}
    \label{tab:ablations}
    \footnotesize\setlength{\tabcolsep}{3pt}
\begin{tabular}{lccc}
\toprule
\textbf{Dataset} & \textbf{UBody} & \textbf{Ego} & \textbf{EHF} \\
 & \textbf{PVE} $\downarrow$ & \textbf{PVE} $\downarrow$ & \textbf{PVE} $\downarrow$ \\
\midrule
\methodname & \textbf{104.4} & \textbf{128.0} & \textbf{95.1} \\
w/o Filtering & 111.1 & 130.7 & 95.4 \\
w/o AMASS~\cite{amass} & 105.3 & 133.0 & 103.5 \\
w/o Hard Mining & 107.9 & 131.2 & 95.4 \\
\bottomrule
\end{tabular}
\end{minipage}
\hfill
\begin{minipage}{0.57\linewidth}
    \centering
    \centering
\caption{\textbf{Domain-Specific Generation.} Replacing 30K of 300K random poses with yoga poses (MOYO) decisively improves MPII Yoga performance with no loss on general benchmarks.}
\label{tab:yoga-poses}
\footnotesize
\setlength{\tabcolsep}{4pt}
\begin{tabular}{lcccc}
\toprule
Dataset & UBody & 3DPW & EHF & MPII Yoga \\
 & PVE $\downarrow$ & PVE $\downarrow$ & PVE $\downarrow$ & PVE $\downarrow$ \\ 
\midrule
270K + 30K yoga & \textbf{97.0} & \textbf{112.7} & 105.3 & \textbf{171.1} \\
300K random & 97.4 & 113.3 & \textbf{104.8} & 199.6 \\
\bottomrule
\end{tabular}
    \label{tab:yoga}
\end{minipage}
\end{table*}

\subsection{Image Quality Evaluation}
A critical advantage of our approach is the substantial improvement in visual realism compared to datasets from traditional rendering engines. \methodname significantly outperforms BEDLAM~\cite{bedlam}, SynBody~\cite{yang2023synbody}, and AGORA~\cite{agora} on both image-quality metrics, improving Inception Score by 76\% over the best baseline (AGORA, $5.55 \rightarrow 9.78$) and reducing FID by 62\% over the best baseline (BEDLAM, $4.55 \rightarrow 1.72$); full metrics in the appendix. This realism reduces the domain gap that limits rendering-engine approaches. To further validate this, we compare models trained on \methodname\ and BEDLAM (\cref{tab:single_datasets}) on two additional in-the-wild benchmarks, MPII~\cite{andriluka14cvpr} and MSCOCO~\cite{mscoco}. As shown in \cref{tab:itw_benchmarks}, the model trained on \methodname\ outperforms BEDLAM on these and on UBody, consistent with the appearance-diversity advantage of generated data on in-the-wild scenarios.


\subsection{Ablation Study}
To evaluate the impact of our pipeline design choices, we compare the downstream model performance on the test set of UBody \cite{ubody}, EgoBody \cite{egobody}, and EHF \cite{smplx}. We use SMPLer-X \cite{smpler}, trained for 6 epochs on datasets generated with different versions of our pipeline, and report the average PVE metric; the results are in \cref{tab:ablations}. Removing any component degrades downstream performance: quality filtering prevents training on misaligned samples, AMASS \cite{amass} adds pose complexity, and hard sample mining ensures learning from the most challenging configurations.

\section{Limitations and Broader Impact}

\methodname\ offers a complementary alternative to rendering engine datasets with distinct trade-offs. While rendering engines provide exact 3D ground truth, our diffusion-based approach offers greater flexibility, scalability, and photorealism, at the cost of reduced 3D-geometric precision. Despite extensive alignment and filtering, generated samples may contain subtle inconsistencies that rendering engines avoid. As our experiments show, this trade-off is benchmark-dependent: generated data benefits in-the-wild appearance variation, rendered data retains 3D precision, and combining them yields the best overall performance. Improving fine anatomical fidelity for hands and facial expressions through region-specific supervision is a promising direction for future work. Our reliance on foundation generative models may also introduce biases from their training data, which could limit pose diversity for underrepresented demographic groups or activities. 
Generating realistic human imagery raises ethical concerns about potential misuse for misleading content or unauthorized representations. Because our pipeline conditions on 3D mesh parameters rather than identity-specific features, the generated images do not depict specific real individuals, reducing privacy risk relative to scraping-based pipelines. Our scalable approach broadens access to high-quality human pose datasets, advancing research in motion analysis, healthcare, and assistive technologies while benefiting researchers who lack access to expensive motion capture systems.
\section{Conclusion}

\methodname\ demonstrates that diffusion-based synthetic data has crossed the utility threshold for human mesh recovery: models trained on our generated data reach performance competitive with much larger rendered datasets, and the two data sources are complementary, generated data benefiting in-the-wild appearance while rendered data retains 3D precision. Our pipeline combines DPO-based control alignment, PNCC-style mesh-to-RGB conditioning, curriculum-based hard sample mining, and multi-stage OKS filtering to maintain 3D-to-2D correspondence at scale, producing a 500K-image dataset whose utility holds across two HMR architectures and several benchmarks. This positions diffusion-generated data as a viable, cost-efficient alternative that scales with compute rather than commissioned artist labor. As diffusion models continue to improve, we expect this approach to scale further in both quality and diversity while maintaining its cost advantage over rendering pipelines.
\newpage
\section*{Acknowledgments}
This work was funded by The Podium Institute for Sports Medicine and Technology, at the University of Oxford. O.K. is supported by a Google unrestricted gift. J.H. acknowledges the generous support of Toyota Motor Europe (TME) and EPSRC (VisualAI, EP/T028572/1). C.R. is supported by ERC StG 101222037-Volute.
{
    \small
    \bibliography{main}
}
\appendix
\section{Additional Analysis}
\subsection{Scaling}
We evaluate PoseDreamer's performance on larger models by training several versions of SMPLer-X-L \cite{smpler}, a ViT-L-based model. The results are shown in Table \ref{tab:scaling}. The model trained on 0.75M samples from two datasets (\methodname and BEDLAM) outperforms the SMPLer-X-L model trained on a mixture of five datasets across all benchmarks, except EgoBody and AGORA (the latter included in the five-dataset train set).

\begin{table}[h]
  \caption{\textbf{Scaling.} We train multiple versions of the SMPLer-X-L and SMPLer-X-S models to demonstrate that \methodname contributes to performance improvements when scaling both data and model size.}
  \label{tab:scaling}
  \centering
  \resizebox{\textwidth}{!}{
  \begin{tabular}{cccc|cccccc}
    \toprule
    
    \#Datasets & \#Inst. & Datasets & \ Backbone &
    AGORA &
    EgoBody &
    UBody & 
    3DPW &
    EHF \\

     & & & & PVE $\downarrow$ & PVE $\downarrow$ & PVE $\downarrow$ & MPJPE $\downarrow$ & PVE $\downarrow$ \\ 

    \midrule

    2 & 0.75M & Ours + BEDLAM (50-50 split) & ViT-S & 156.8 & \textbf{106.1} & \textbf{97.6} & \textbf{96.8} & \textbf{84.3} \\
    5 & 0.75M & SMPLer-X-S 5 datasets & ViT-S & \textcolor{gray}{119.0} & 114.2 & 110.1 & 110.2 & 100.5 \\
        
    \midrule
    2 & 0.75M & Ours + BEDLAM (500K-250K) & ViT-L &
    138.7 & 101.1 & \textbf{93.8} & \textbf{88.5} & \textbf{73.3} \\
    
    2 & 0.75M & Ours + BEDLAM (50-50 split) & ViT-L &
    138.1 & 114.2 & 107.4 & 100.1 & 93.9 \\

    5 & 0.75M & SMPLer-X-L 5 datasets & ViT-L &
    \textcolor{gray}{88.3} & \textbf{98.7} & 110.8 & 97.8 & 89.5 \\
    
    \midrule

    2 & 1.5M & Ours + BEDLAM (50-50 split) & ViT-L & 
    131.3 & 99.6 & \textbf{94.6} & 85.6 & 67.5 \\
    
    10 & 1.5M & SMPLer-X-L 10 datasets & ViT-L & 
    \textcolor{gray}{82.6} & \textcolor{gray}{69.7} & 104.0 & \textbf{82.5} & \textbf{64.0} \\
    
    \bottomrule
    \end{tabular}}
\end{table}

Notably, the model trained on an equal split of \methodname and BEDLAM underperforms compared to the model trained on 250K samples from BEDLAM and 500K samples from \methodname. This indicates that \methodname's diversity and photorealism continue to improve performance at scale, while BEDLAM's contribution appears to saturate.

We also compare the SMPLer-X-L model trained on 1.5M samples from two datasets (\methodname and BEDLAM) with a version trained on a mixture of ten datasets. The results are comparable across all benchmarks not included in the ten-dataset training set, that is, every benchmark except AGORA and EgoBody. Moreover, the model trained on 2 datasets outperforms the one trained on 10 datasets on the in-the-wild UBody benchmark.

These results demonstrate that our dataset provides substantial diversity in poses and environments, leading to improved model performance when scaling both model size and the number of training samples.

\subsection{Comparison with Diffusion-HPC on SMART}
\begin{figure*}[h!]
\centering
\includegraphics[width=0.99\textwidth]{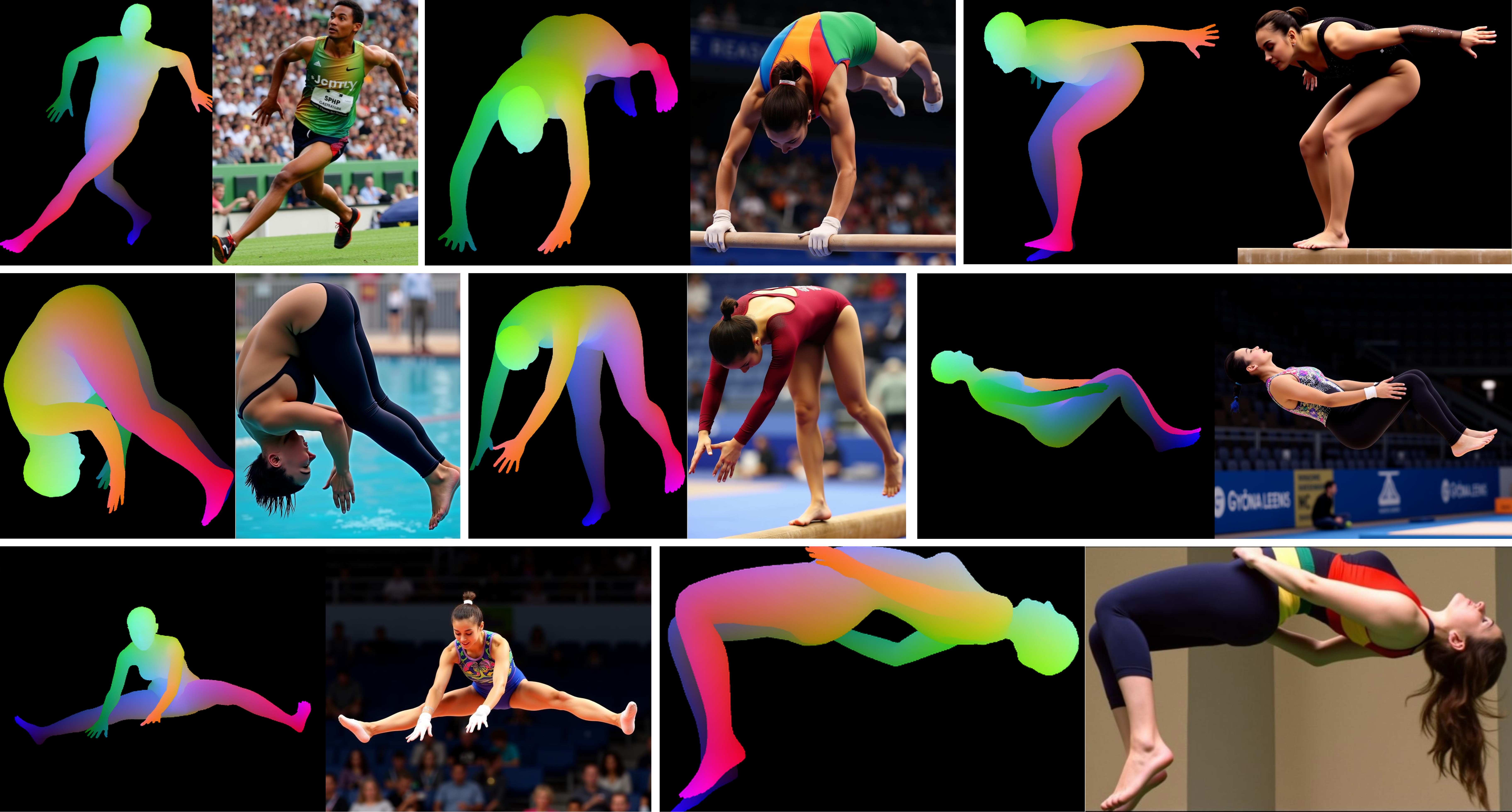}
\caption{\textbf{SMART Poses Generated.} Examples of images generated using poses from the SMART dataset.}
\label{fig:smart-gen}
\end{figure*}

To complement the yoga experiment in the main paper, we compare \methodname\ against Diffusion-HPC~\cite{diffusion-hpc} on the SMART dataset~\cite{smart} of challenging athletic poses. We follow the experimental setup reported in their paper: starting from approximately 100 real images per sport, we generate 3$\times$ augmented synthetic images and use them to finetune the downstream regressor, comparing against the numbers reported by Diffusion-HPC. For the SMART experiments, Diffusion-HPC's released setup initializes generation from real target-domain images, so its outputs retain elements of real-image appearance. \methodname, by contrast, generates from scratch conditioned only on pose, with no exposure to the target distribution. We further apply our OKS-based filtering, retaining only generations with an OKS score above 0.8, so the downstream model is trained on fewer but higher-quality images. As shown in Table~\ref{tab:diffusion-hpc}, \methodname\ matches or exceeds Diffusion-HPC on four of six actions despite this difference in setup. Figure~\ref{fig:smart-gen} shows examples of generations from SMART poses, illustrating that our pipeline handles the extreme articulations and unusual viewpoints typical of athletic motion.

\begin{table}[h]
\centering
\caption{\textbf{SMART Comparison.} Per-action PCK ($\uparrow$) against Diffusion-HPC across six athletic actions. ``Extra Cond.''\ indicates whether the method requires additional conditioning beyond pose: Diffusion-HPC uses real SMART images as image-to-image starting points, while \methodname\ generates from scratch conditioned only on pose.}
\resizebox{\linewidth}{!}{%
\begin{tabular}{lccccccc}
\toprule
\textbf{Method} & \textbf{Extra Cond.} & \textbf{Diving} & \textbf{Pole Vault} & \textbf{High Jump} & \textbf{Uneven Bars} & \textbf{Balance Beam} & \textbf{Vault} \\
\midrule
Diffusion-HPC & \checkmark & 79.2 & \textbf{77.7} & 78.1 & 44.1 & 85.1 & \textbf{66.9} \\
\methodname & $\times$ & \textbf{81.7} & 73.7 & \textbf{81.7} & \textbf{45.1} & \textbf{85.4} & 65.1 \\
\bottomrule
\end{tabular}%
}

\label{tab:diffusion-hpc}
\end{table}

\subsection{Spatial Control Ablations}
To train a reliable control mechanism for dataset sample generation, we experimented with different representations of the control image. Our initial attempt involved generating image–label pairs by running a pretrained DensePose \cite{densepose} model on a subset of LAION \cite{laion} images. However, these predictions were often inaccurate or incomplete, or had low-quality boundaries, resulting in poor pixel alignment for our control mechanism. 

We next considered rendering DensePoseCOCO \cite{densepose} and AGORA \cite{agora} ground truth annotations to ensure high-quality image pairs. Initially, we rendered SMPL-X parameters using the colormap from Continuous Surface Embeddings \cite{cse}. This colormap, however, maps each vertex to a value between 0 and 1, which provides insufficient visual variation across poses, particularly for distinguishing head orientations or body configurations. Moreover, this colormap is only defined for SMPL parameters, whereas our goal is to model the full body pose using the SMPL-X model \cite{smplx}. We were able to obtain a stable and reliable control using the PNCC colormap \cite{pncc}, and mapping each X,Y,Z normalized coordinate to R,G,B values.  Following the ablations from the main paper, we compare different representation choices for the conditioning image and report their impact on downstream model performance in Table \ref{tab:control_type_ablations}. Using high-quality image pairs and a more detailed colormap significantly improves downstream model performance. 

\begin{table}[h]
\centering
\caption{\textbf{Control Representations.} We compare the effect of different representation choices for the control model training. The comparison is based on downstream model performance. }
\label{tab:control_type_ablations}
\resizebox{0.6\linewidth}{!}{
\begin{tabular}{lccc}
\toprule
Control 2D Representation & UBody & Egobody & EHF \\
 & PVE $\downarrow$ & PVE $\downarrow$ & PVE $\downarrow$ \\
\midrule
\methodname (PNCC) & \textbf{104.43} & \textbf{128.02} & \textbf{95.08} \\
CSE \cite{cse} colormap & 129.83 & 130.66 & 130.71 \\
Densepose predictions  & 119.95 & 153.10 & 151.40 \\
\bottomrule
\end{tabular}
}
\end{table}

\subsection{Filtering Statistics}
\begin{figure*}[p] 
  \centering
  \includegraphics[width=\textwidth, keepaspectratio]{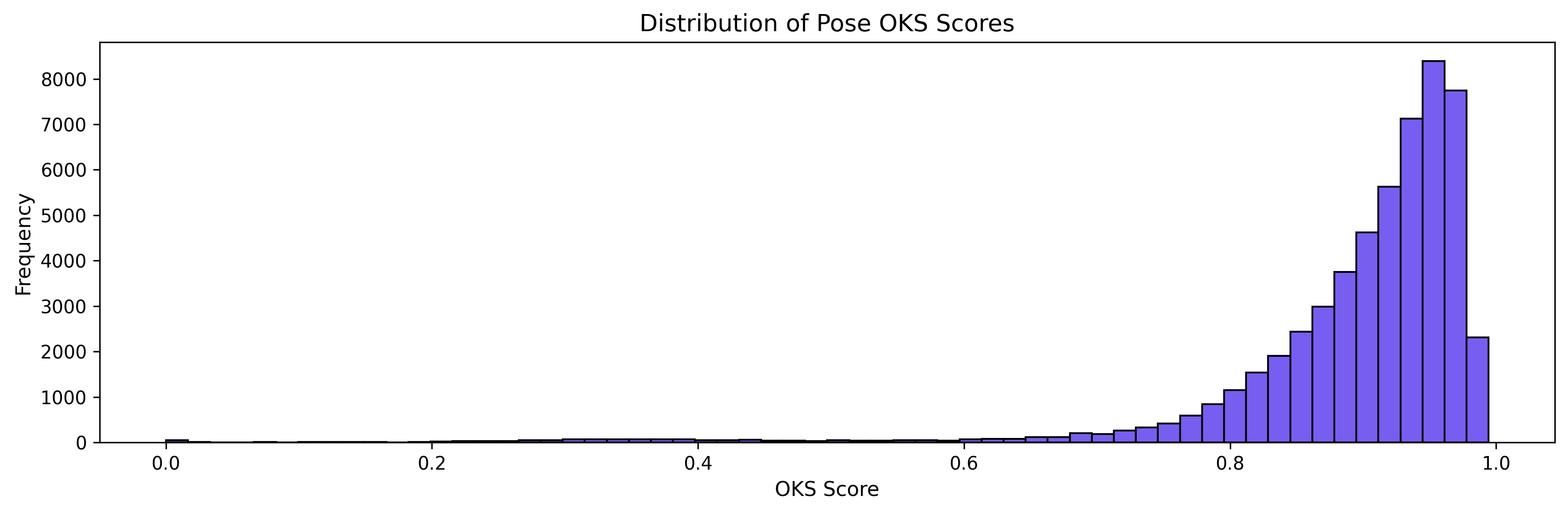}\\[6pt]
  \includegraphics[width=\textwidth, keepaspectratio]{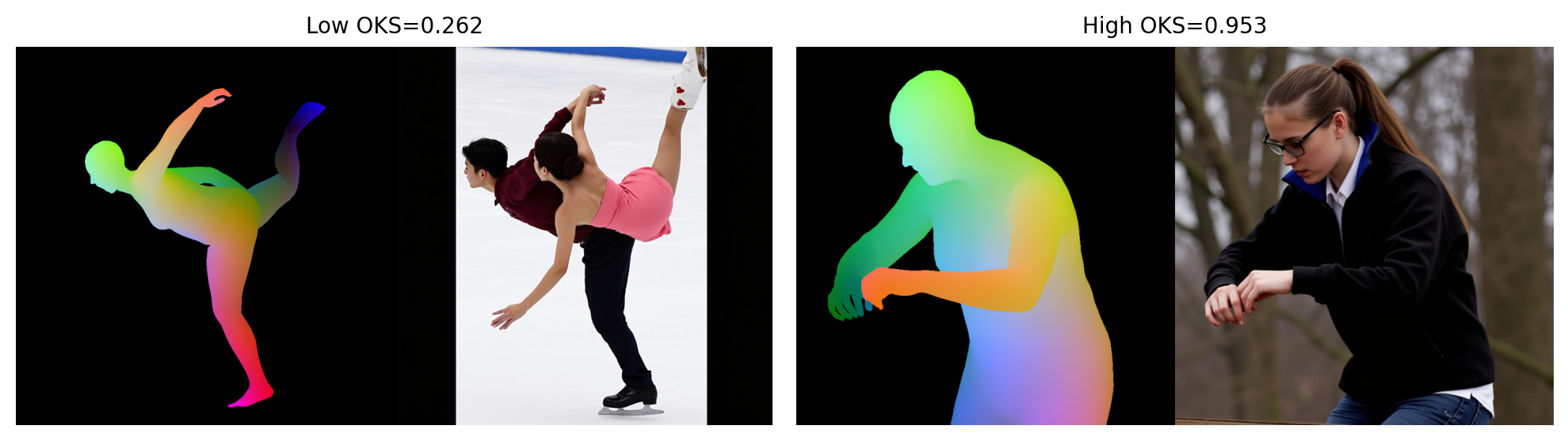}\\[6pt]
  \includegraphics[width=\textwidth, keepaspectratio]{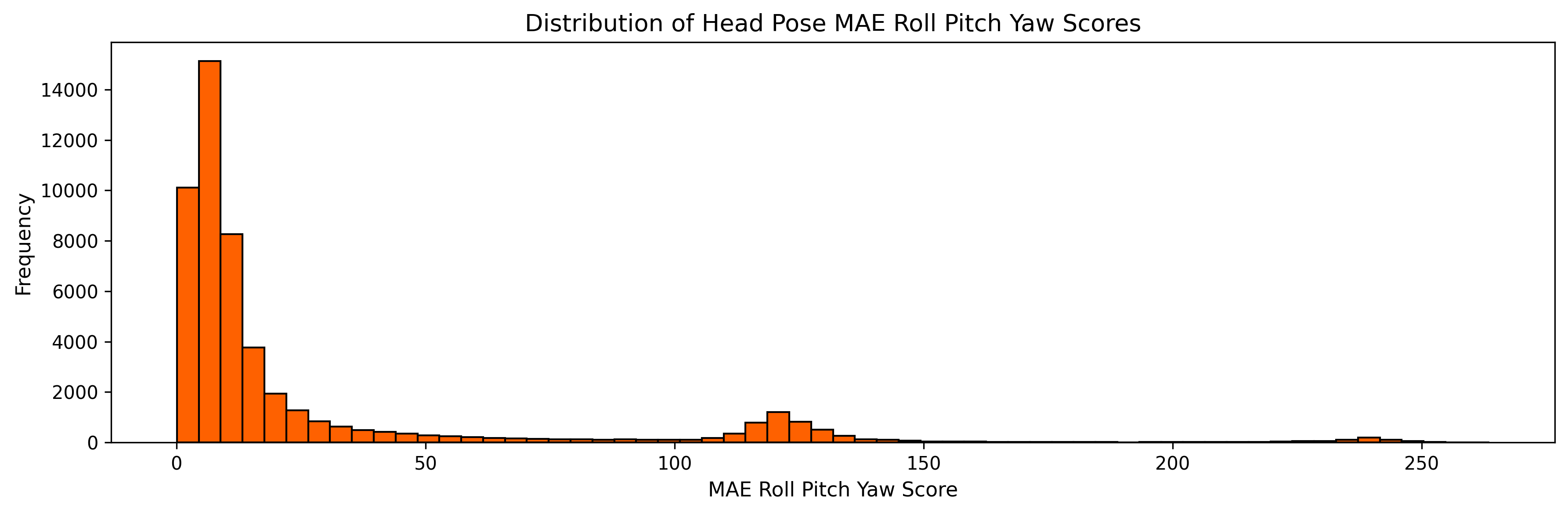}\\[6pt]
  \includegraphics[width=\textwidth, keepaspectratio]{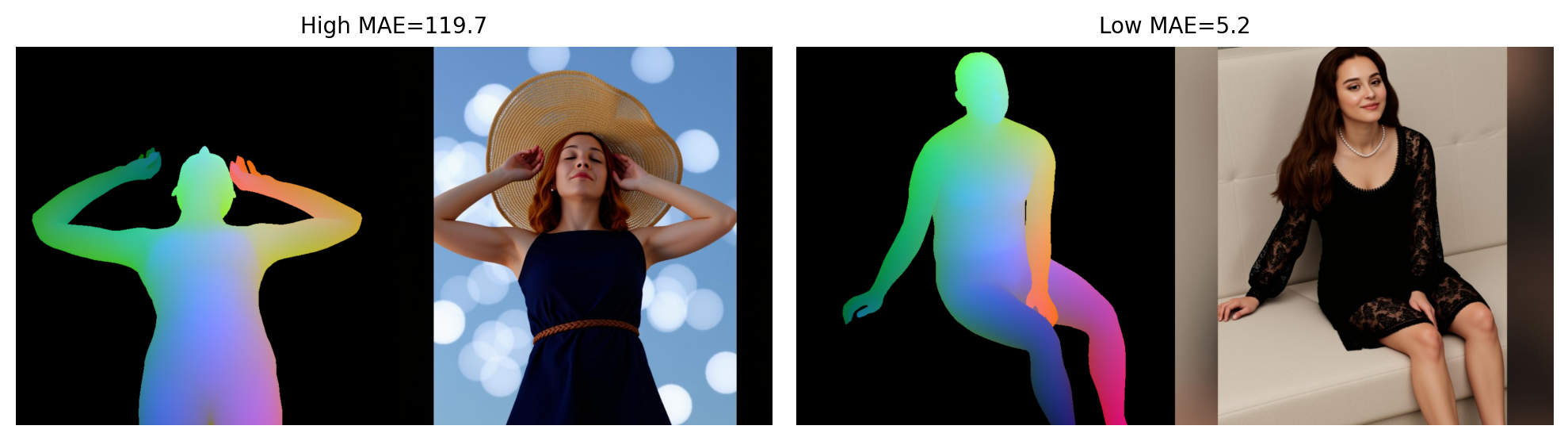}
  \caption{\textbf{Filtering statistics.} Distribution of the pose OKS errors, and head pose MAE for roll pitch yaw, together with examples of low- and high-scoring images.}
  \label{fig:filtering_stats}
\end{figure*}

To ensure the quality of the generated dataset, we introduce a filtering pipeline. We provide additional statistics on the scores of the generated images. Figure \ref{fig:filtering_stats} shows the histogram of the OKS score distribution for pose predictions on 60,000 generated images, along with examples of low- and high-scoring images. In addition, we present the histogram of the distribution of mean absolute errors (MAE) for roll, pitch, and yaw in the 3D head pose parameter predictions. The thresholds used for the final dataset selection are 0.8 for the OKS score and 25° for the head pose error.

\begin{figure}[t]
\centering
\includegraphics[width=0.99\linewidth]{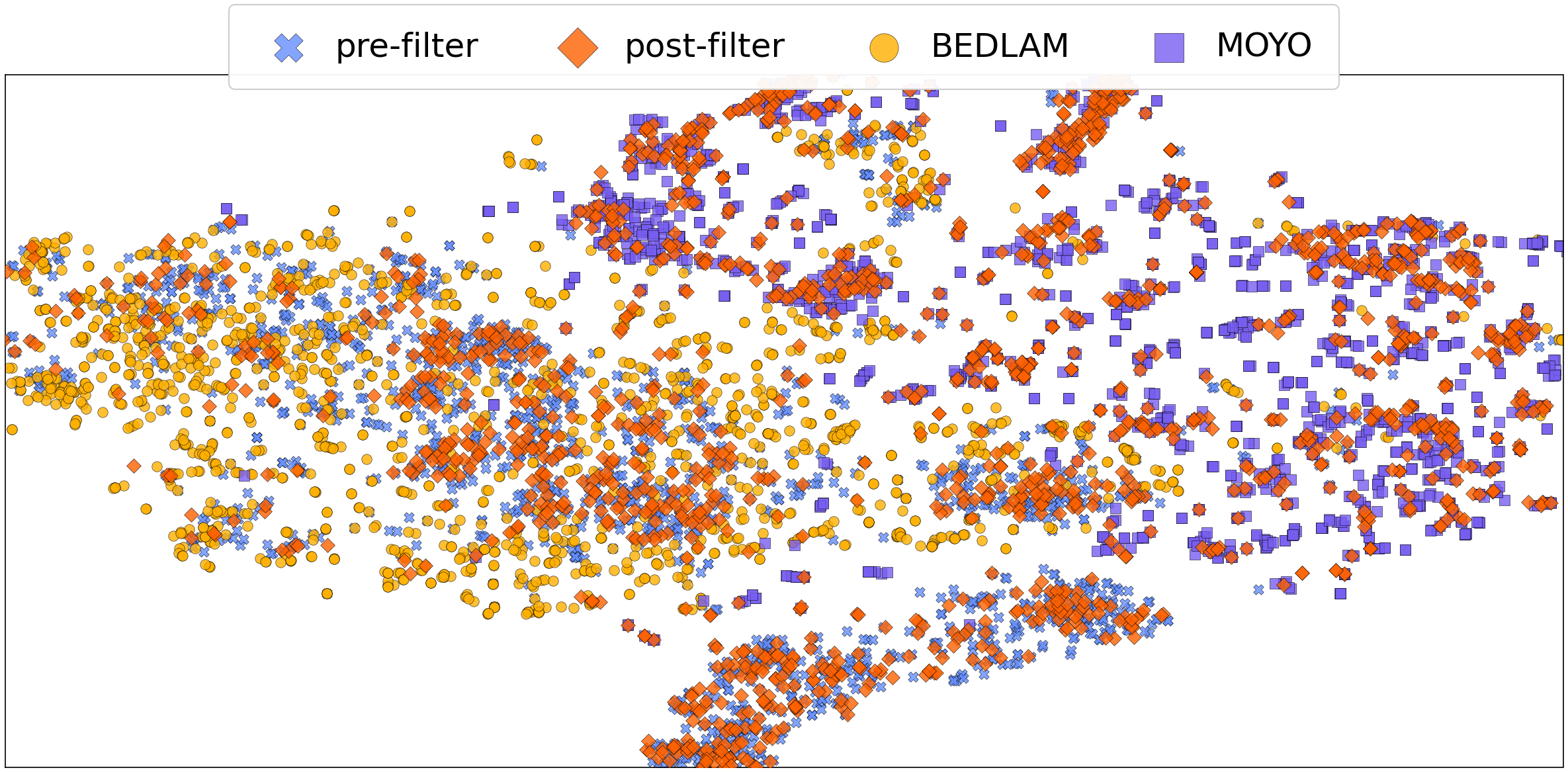}
\caption{\textbf{Pose distribution.} Filtering preserves the overall pose space; BEDLAM and MOYO occupy subregions of \methodname's distribution.}
\label{fig:pose_tsne}
\end{figure}

\paragraph{Pose distribution and pass rates.}
To verify that filtering does not systematically discard valid-but-hard poses, we compare pre- and post-filter pose distributions in \Cref{fig:pose_tsne} (t-SNE over SMPL-X body-pose parameters). Filtering preserves the overall pose space, confirmed by the low JS divergence between pre- and post-filter distributions (0.031 for LAION+AMASS, 0.042 for MOYO), while BEDLAM and MOYO occupy subregions of our broader distribution. Pass rates differ by source; for instance, the OKS filter pass rate is 77.7\% for LAION+AMASS and 60.1\% for MOYO, consistent with MOYO's out-of-domain yoga poses (neither the control model nor DPO was trained on yoga).

\paragraph{Threshold sensitivity.}
Sweeping OKS (0.7--0.9) and head-pose MAE (15$^\circ$--25$^\circ$) thresholds over 400K generated images changes downstream performance by at most $\sim$3\%. Best values depend on training setup and evaluation set; we use 0.8 (OKS) and 25$^\circ$ (MAE) as a balanced overall recipe.

\paragraph{Distribution balancing.}
We balance gender and ethnicity when sampling identity descriptors during generation. Other distributions (e.g., lighting or activity mix) are task-dependent and can be adjusted through prompting without retraining the pipeline.

\subsection{Image Quality}
\begin{figure*}[p] 
  \centering

  \newcommand{\panelHeight}{0.28\textheight} 

  \begin{minipage}[b]{0.49\textwidth}
    \centering
    \includegraphics[width=\linewidth,height=0.33\textheight,keepaspectratio]{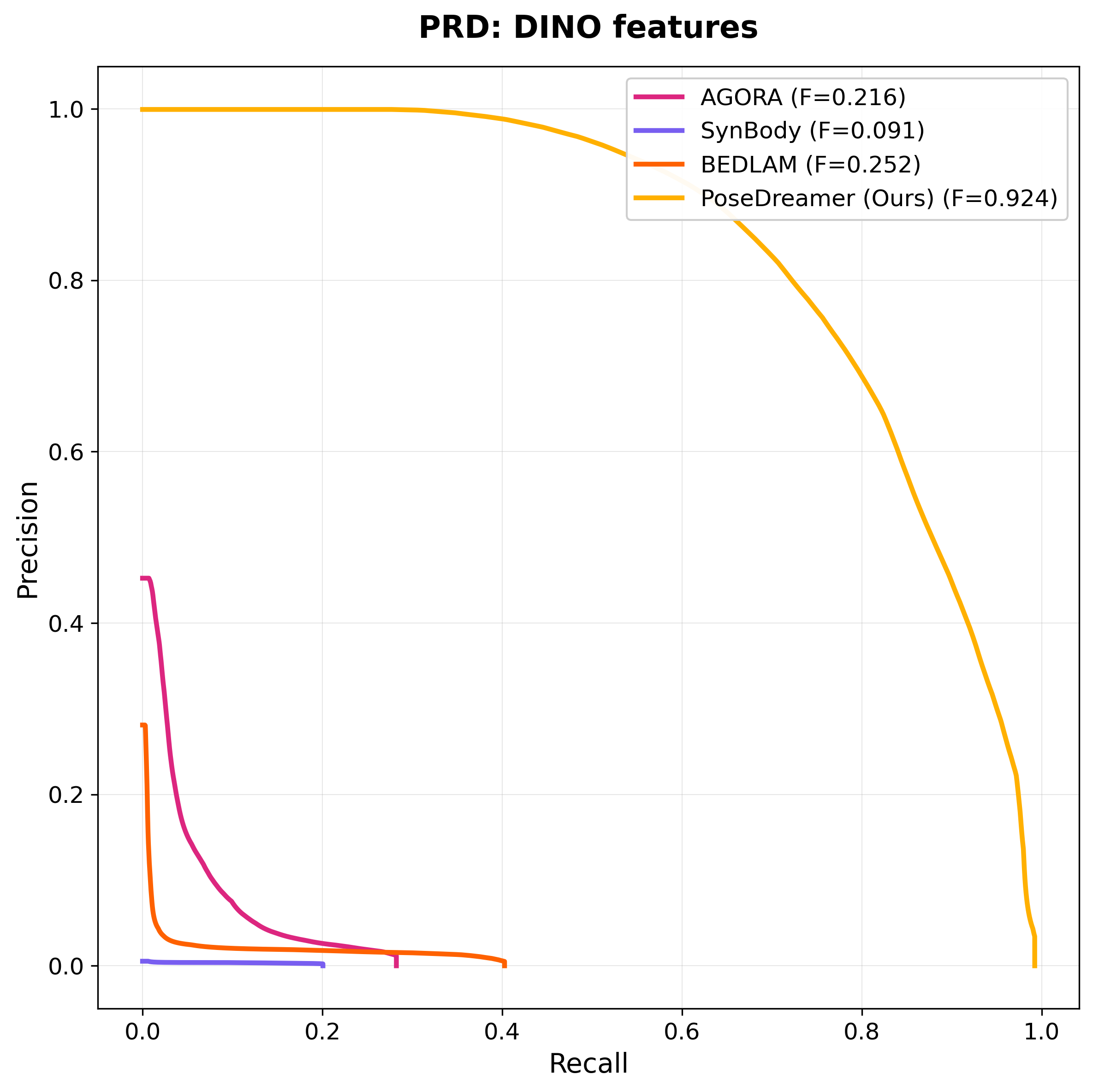}
  \end{minipage}\hfill
  \begin{minipage}[b]{0.49\textwidth}
    \centering
    \includegraphics[width=\linewidth,height=0.33\textheight,keepaspectratio]{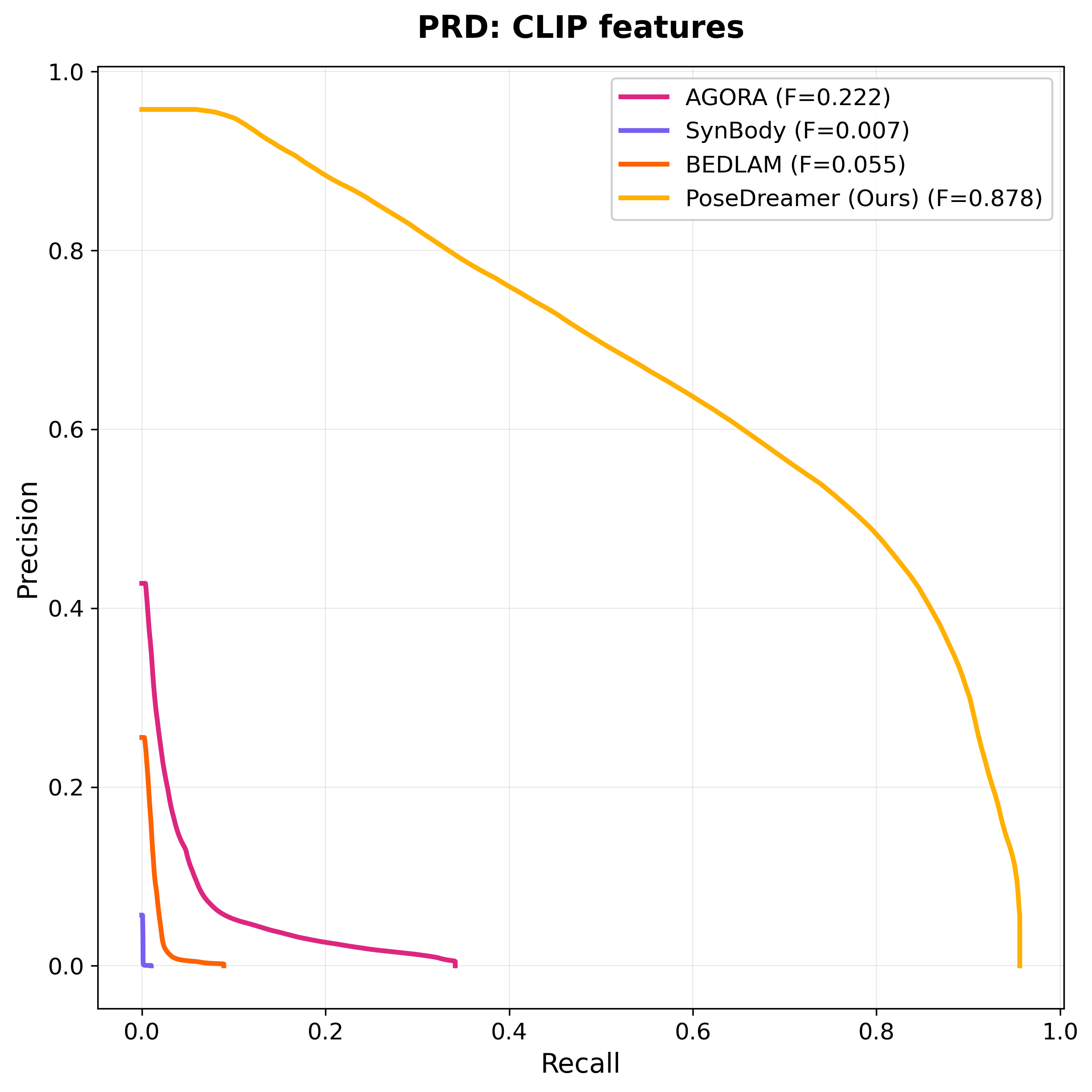}
  \end{minipage}
  
  \vspace{6pt} 

  \begin{minipage}[b]{0.49\textwidth}
    \centering
    \includegraphics[width=\linewidth,height=\panelHeight,keepaspectratio]{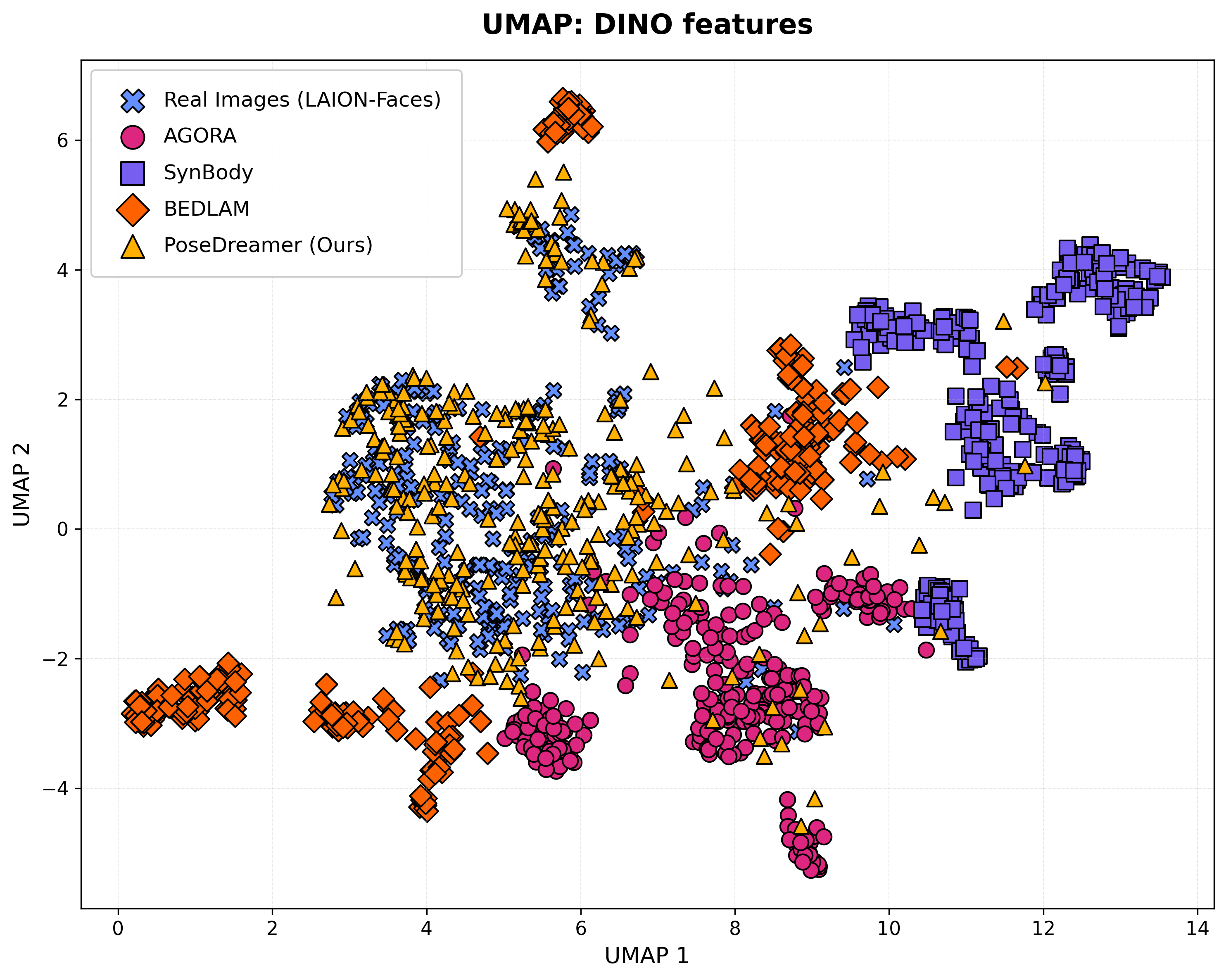}
  \end{minipage}\hfill
  \begin{minipage}[b]{0.49\textwidth}
    \centering
    \includegraphics[width=\linewidth,height=\panelHeight,keepaspectratio]{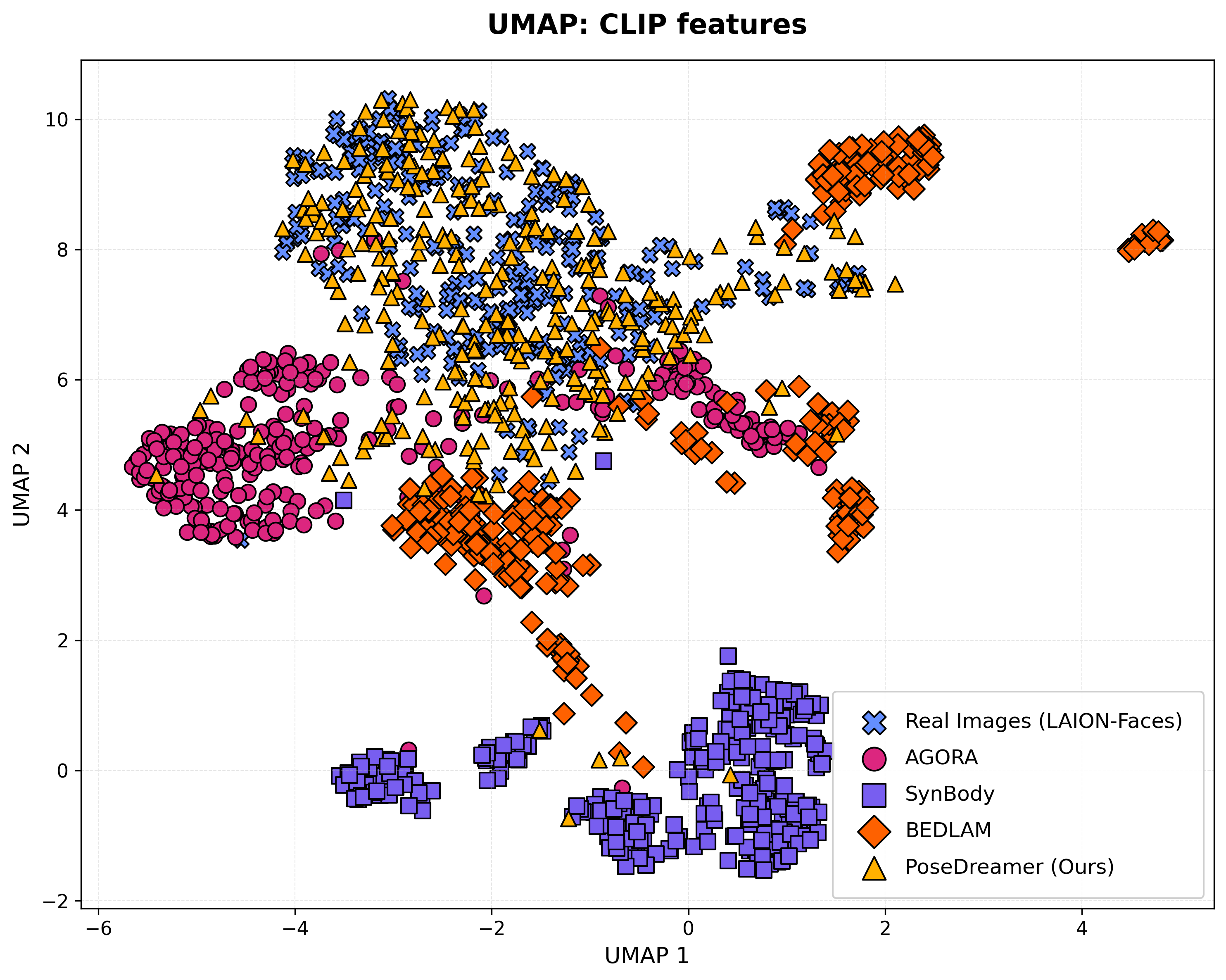}
  \end{minipage}
   
  \vspace{6pt}

  \begin{minipage}[b]{0.49\textwidth}
    \centering
    \includegraphics[width=\linewidth,height=\panelHeight,keepaspectratio]{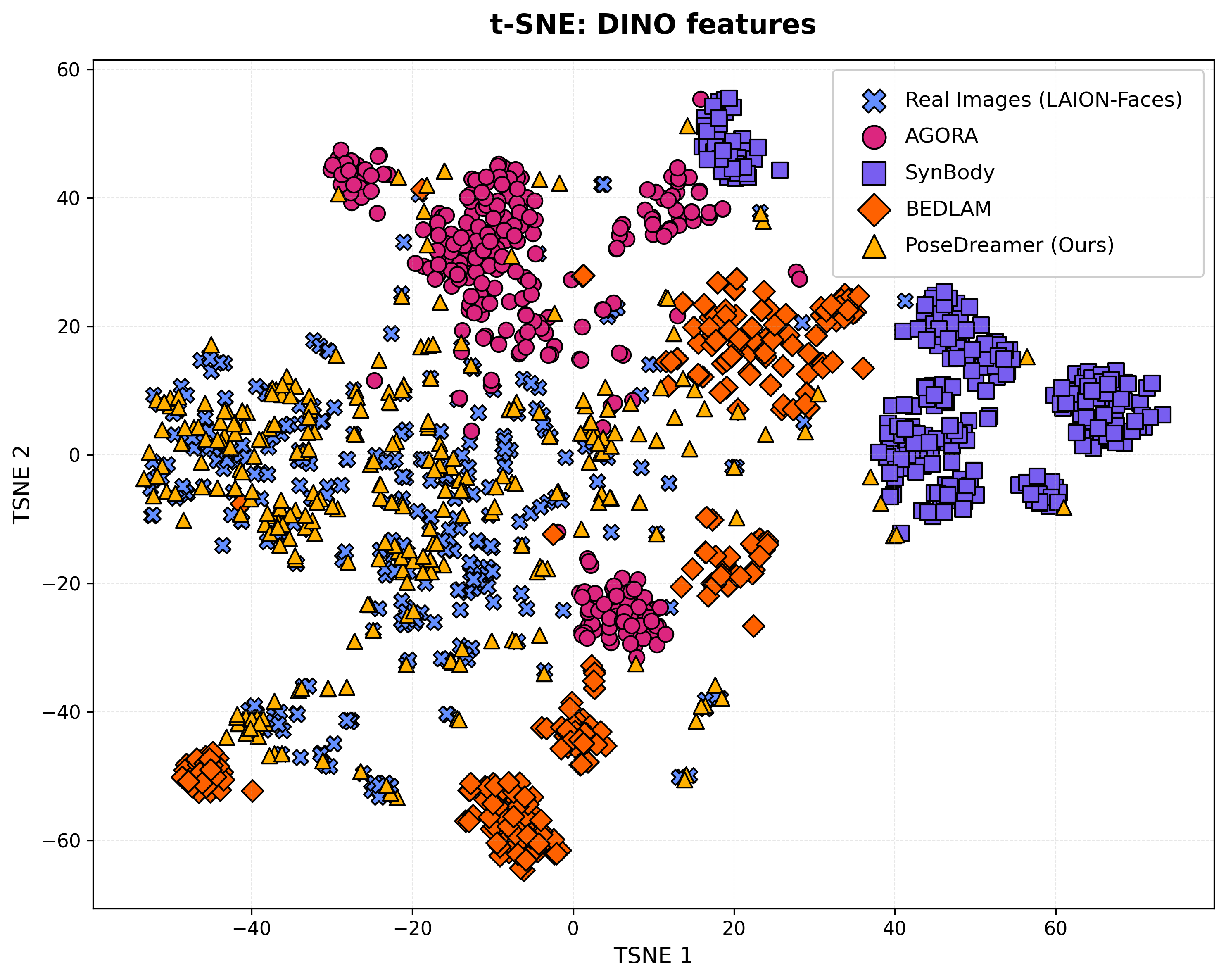}
  \end{minipage}\hfill
  \begin{minipage}[b]{0.49\textwidth}
    \centering
    \includegraphics[width=\linewidth,height=\panelHeight,keepaspectratio]{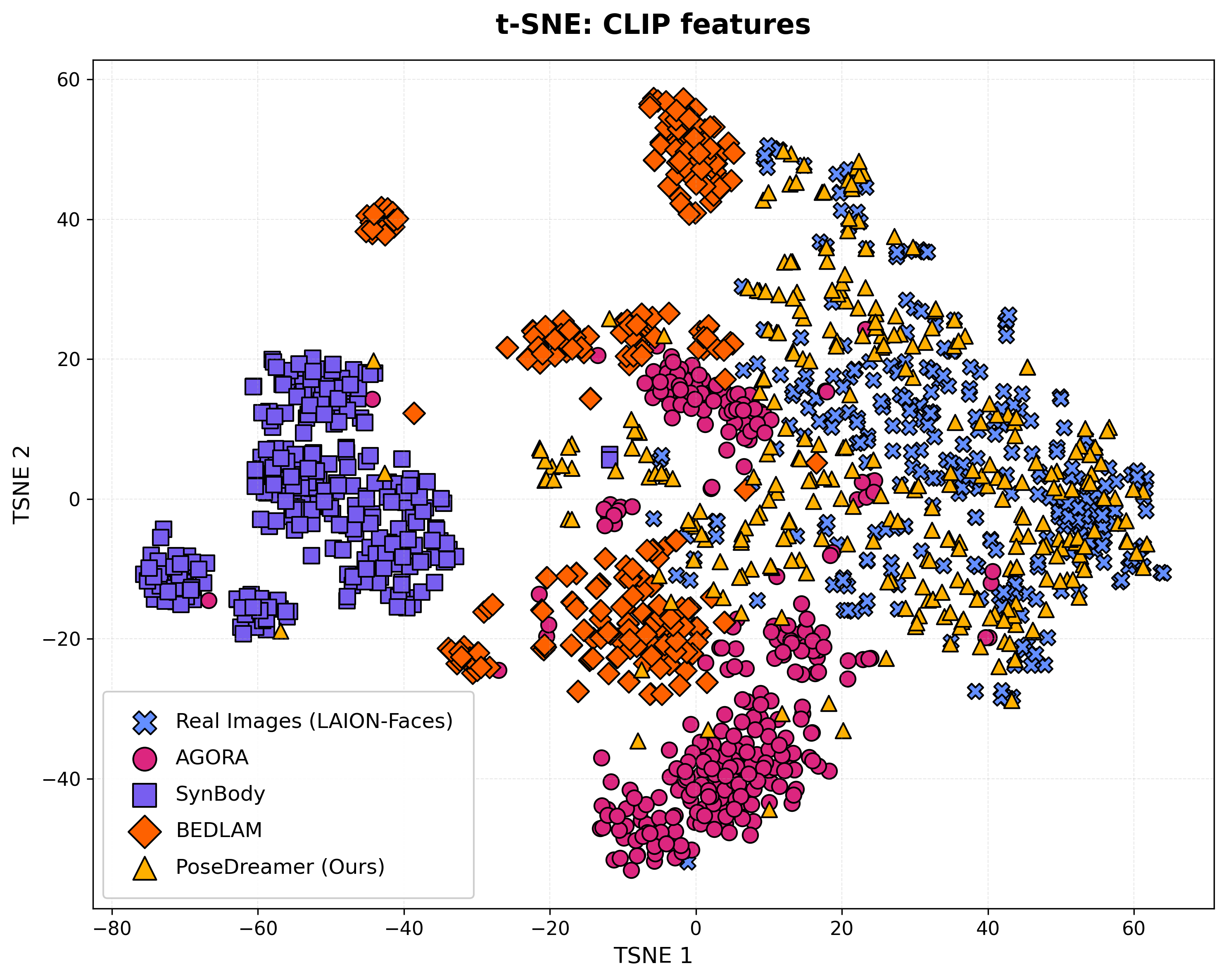}
  \end{minipage}

  \vspace{4pt}
  \caption{\textbf{Image Quality Analysis.} Results for quality and coverage of the samples following \cite{prd}. UMAP \cite{umap} and TSNE \cite{tsne} projections of DINO \cite{dinov3} and CLIP \cite{clip} image features for \methodname compared to real images from the LAION faces dataset and other synthetic datasets. Samples from our generated dataset more closely follow the distribution of the real dataset. }
  \label{fig:supp_image_quality}
\end{figure*}

To quantitatively evaluate the synthetic-to-real domain gap, we compute the quality and coverage of samples produced by a generative model following \cite{prd}, and compare them to the faces subset of the LAION dataset \cite{laion}. The results are shown in Figure \ref{fig:supp_image_quality}. We observe that the generated images closely match the real distribution, whereas rendered datasets achieve a much lower F-score.
Furthermore, UMAP \cite{umap} and t-SNE \cite{tsne} projections of DINO-v3 \cite{dinov3} and CLIP \cite{clip} image features show that \methodname\ samples occupy the same region of the real data manifold. In comparison, BEDLAM \cite{bedlam}, AGORA \cite{agora}, and SynBody \cite{yang2023synbody} form isolated clusters. \cref{tab:image_quality_supp} shows the corresponding Inception Score and FID metrics: \methodname\ substantially outperforms rendered synthetic datasets on both.

\begin{table}[htb]
\centering
\footnotesize
\caption{\textbf{Image Quality:} Diffusion-generated data achieves superior realism than rendering-based synthetic datasets.}
\begin{tabular}{lcc}
\toprule
\textbf{Dataset} & \textbf{Inception Score} $\uparrow$ & \textbf{FID} $\downarrow$ \\
\midrule
\methodname & \textbf{9.78} & \textbf{1.72} \\
BEDLAM \cite{bedlam} & 4.35 & 4.55 \\
SynBody \cite{yang2023synbody} & 5.39 & 5.19 \\
AGORA \cite{agora} & 5.55 & 8.14 \\
\bottomrule
\end{tabular}

\label{tab:image_quality_supp}
\end{table}

\subsection{2D Reprojection}
Our synthetic dataset's precise ground truth annotations enable the incorporation of additional explicit supervision signals within the SMPL-X topology. We explore how rich synthetic data can improve mesh recovery through enhanced architectural supervision.
We introduce a minor modification to the SMPLer-X architecture by adding intermediate visual marker predictions. While SMPLer-X originally predicts 3D joint locations as input to pose parameter heads, we extend this concept by incorporating visual markers as a sparse set of 64 fixed vertices on the mesh surface, following VirtualMarker~\cite{virtualmarker}. Unlike joints, which primarily capture pose information, visual markers are more sensitive to shape variations, providing valuable signals for predicting shape parameters.

This approach provides shape-sensitive supervision while maintaining efficiency through sparse representation.
Our modified architecture predicts both joints and visual markers simultaneously, feeding marker predictions to the shape parameter head as additional input. We apply supervision losses to markers in both 3D and 2D spaces, directly on predicted markers and after reconstructing them from predicted avatars, analogous to joint supervision in the original architecture.
\Cref{tab:2d_reproj} demonstrates the effectiveness of this enhanced supervision. Our model with visual marker supervision achieves 2D reprojection errors approaching SMPLer-X trained on 32 combined real and synthetic datasets, demonstrating that high-quality synthetic supervision can achieve competitive performance with significantly fewer data sources.

\begin{table}[h!]
\centering
\caption{\textbf{2D Reprojection Error Analysis.} \methodname provides rich supervision via visual markers, achieving 2D reprojection accuracy approaching that of models trained on 32 datasets while using only our single synthetic source.}
\label{tab:2d_reproj}
\resizebox{0.7\linewidth}{!}{
\begin{tabular}{lccc}
\toprule
Training Data & EHF & EgoBody & UBody \\
& \small{2D Reproj $\downarrow$} & \small{2D Reproj $\downarrow$} & \small{2D Reproj $\downarrow$} \\
\midrule
32 Datasets & \textbf{0.013} & 0.026 & \textbf{0.024} \\
\methodname & 0.017 & 0.027 & 0.030 \\
\methodname + Markers & 0.015 & \textbf{0.024} & 0.025 \\
\bottomrule
\end{tabular}
}
\end{table}

\begin{figure*}[h!]
\includegraphics[width=0.99\textwidth]{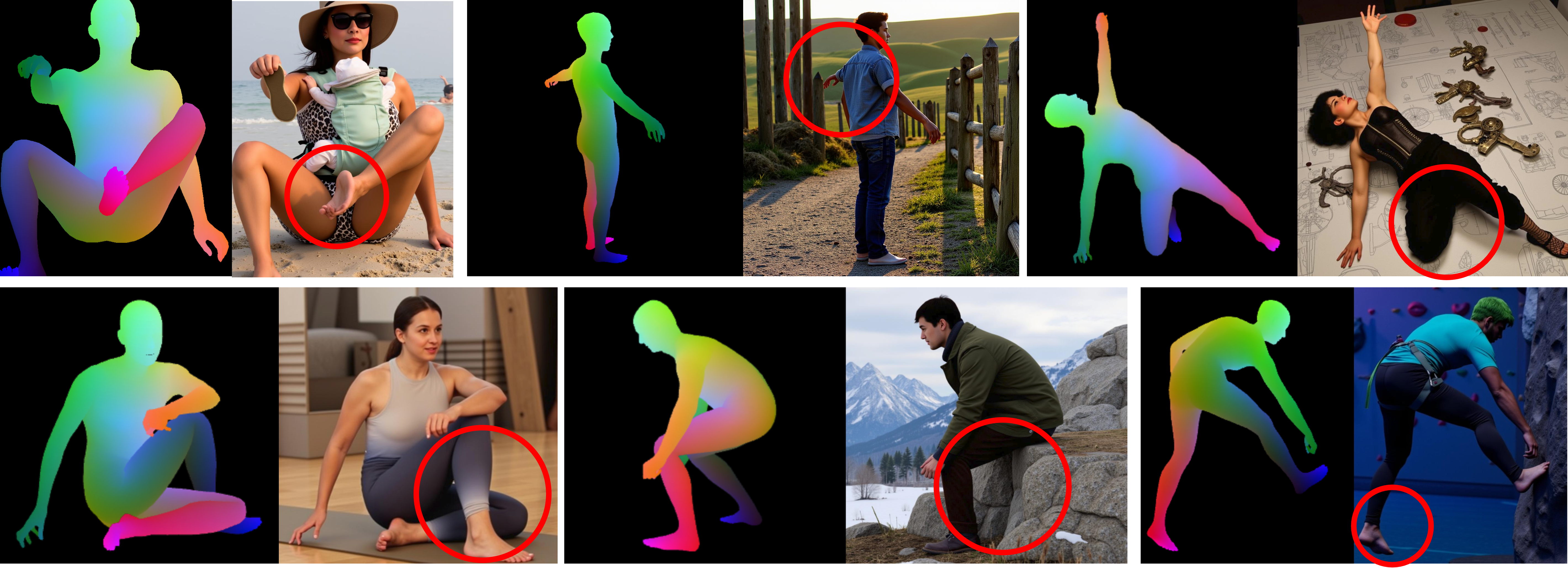}
\centering
\caption{\textbf{\methodname Failure cases.} There are some failure cases for our method. These include unrealistic body shapes for complex input poses or input descriptions that involve complex interactions with other people. Additionally, it can happen that the model fails to understand the 3D pose perspective of the input, or that it places the person in an unrealistic scenario (for example, floating with no support).}
\label{fig:failure_cases}
\end{figure*}
\section{Failure Cases}

While our method is effective, it has limitations: the generated images are not always fully photorealistic or perfectly aligned with the control input. Most mistakes are removed during filtering, but minor ones occasionally slip through. Manual inspection of 100 random samples found 3 hand-detail artifacts (3\%) and no face artifacts. These match known FLUX limitations rather than DPO-induced failures: DPO's OKS-based reward has blind spots, as anatomical errors that preserve keypoint positions do not affect OKS and pass through. Region-specific hand and face losses are therefore a natural future direction.

The failures that do survive fall into three groups. Complex body poses, such as yoga positions or intricate multi-person interactions described in the text, can produce deformed body parts or additional limbs. Physically implausible scenarios arise, such as people appearing to float. Finally, the model sometimes struggles with the 3D-to-2D mapping of limb joints, bending legs in the generated image that are straight in the control input as it incorrectly adjusts them for image perspective. Examples are shown in Figure~\ref{fig:failure_cases}.

\begin{figure}[h!]
    \centering
\begin{tikzpicture}[font=\scriptsize] 

\node (A) at (0,0) 
{\includegraphics[width=0.6\linewidth]{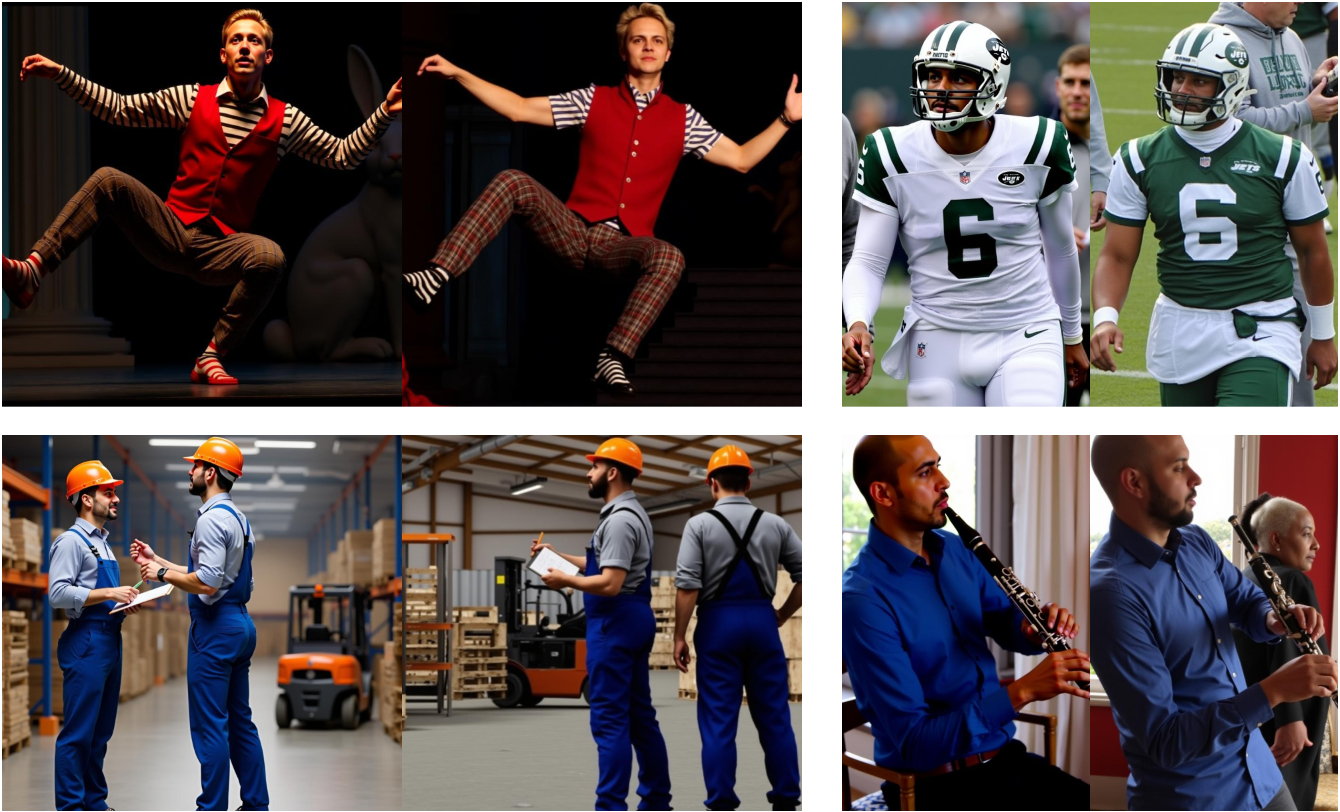}};

\node (B) [anchor=west] 
at ([xshift=-3.4cm,yshift=0.3cm] A.north) {\tiny FLUX.1-Dev};

\node (C) [anchor=west] 
at ([xshift=-1.4cm,yshift=0.3cm] A.north) {\tiny +Realism LoRA};

\node (D) [anchor=west] 
at ([xshift=0.65cm,yshift=0.3cm] A.north) {\tiny FLUX.1-Dev};

\node (E) [anchor=west] 
at ([xshift=2.15cm,yshift=0.3cm] A.north) {\tiny +Realism LoRA};

\node (F) [anchor=north] 
at ([yshift=0.1cm] A.south) 
{\includegraphics[width=0.6\linewidth]{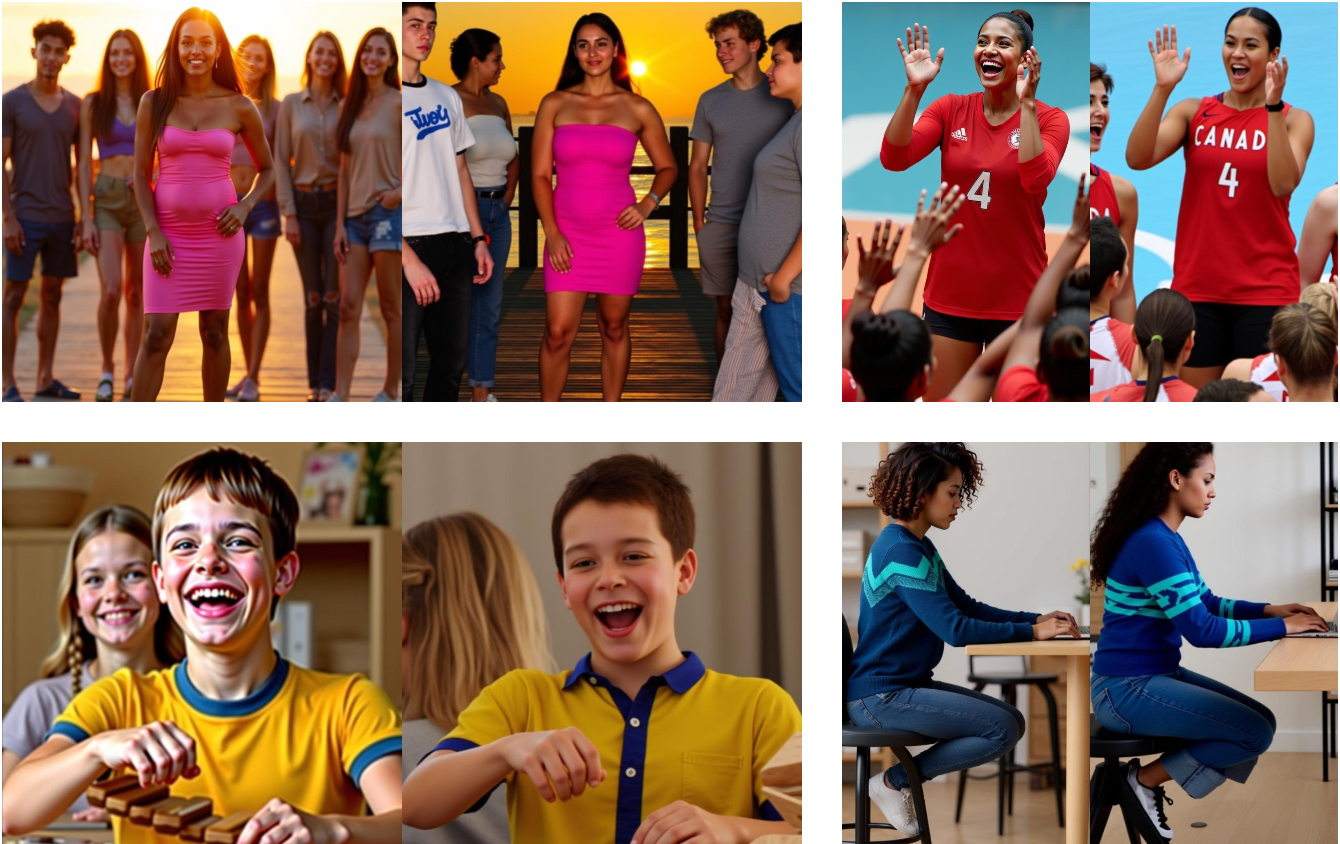}};

\end{tikzpicture}

\caption{\textbf{Effectiveness of Realism Adapter.} We show comparisons of the same image generated without (left) and with the realism LoRA adapter. Images generated without it tend to be more oversaturated, giving an "AI look".}
\label{fig:realism_lora}
\end{figure}

\section{Mitigating Distribution Gap in Diffusion-Generated Data}
Diffusion models trained on large-scale web datasets like LAION often exhibit aesthetic biases, resulting in oversaturated, heavily filtered imagery with an artificial "AI look." This happens because these models learn to copy the visual styles common in retouched internet images, which are quite different from natural image distributions, leading the generated synthetic data to look unrealistic.

To address this distribution gap, we use a specialized LoRA adapter~\cite{lora} in our pipeline, trained on real-world images to reduce aesthetic artifacts from foundation models. This adapter learns to transform images from the biased aesthetic style of web-trained diffusion models toward more natural image appearance. The modular design of EasyControl~\cite{easycontrol} allows to combine the aesthetic/style adapter with our spatial control LoRA at inference time, producing images that maintain both precise spatial consistency with control inputs and improved realistic appearance.

\Cref{fig:realism_lora} demonstrates the effectiveness of this approach, showing substantial reduction in oversaturation and artificial enhancement artifacts. This aesthetic correction represents a crucial step in bridging the domain gap between synthetic training data and real-world deployment scenarios.

\section{Control Model Alignment: Implementation Details}

For DPO training, we generate 40,000 preference pairs by creating 4 image variants for each caption and 3D mesh combination, then ranking them based on OKS scores to establish preferred and less preferred samples. We use a lower $\beta$ value of 250 compared to the original Flow-DPO \cite{videoreward} value of 500, as we find that control models require less regularization due to their more constrained generation space. Additionally, traditional ControlNet approaches may face challenges with RL-based alignment, as both the control mechanism and alignment procedures operate on diffusion model activations, potentially leading to optimization conflicts that can compromise training stability. In contrast, EasyControl \cite{easycontrol} modular adapter architecture we use provides cleaner separation between control and alignment components, facilitating more stable joint optimization. 
\section{Additional Visualizations}
\begin{figure*}[h!]
\includegraphics[width=0.99\textwidth]{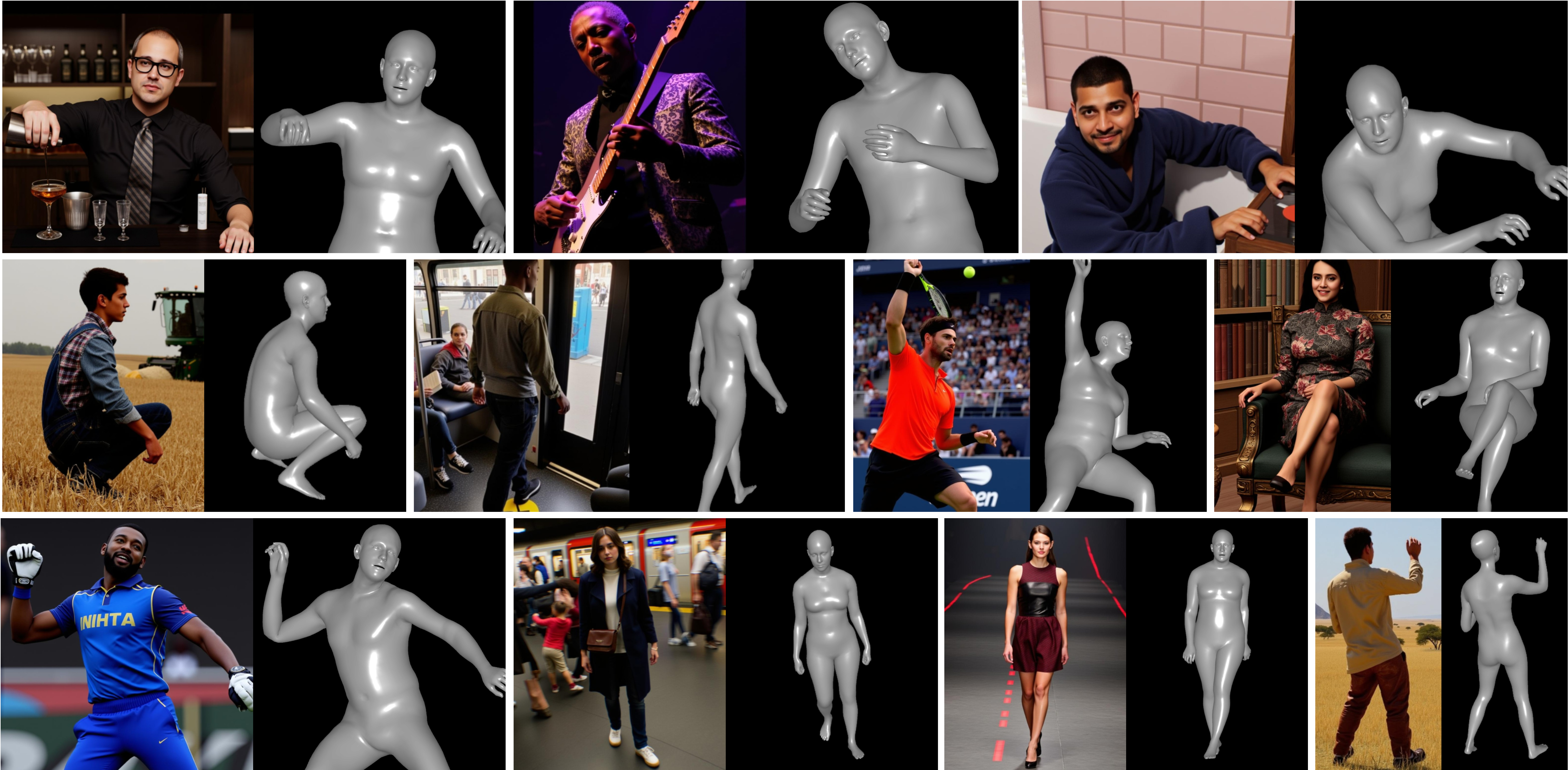}
\centering
\caption{\textbf{\methodname Dataset Samples.} Examples from \methodname dataset showing diverse poses, scenes, and high-quality image generation with precise 3D mesh correspondence.}
\label{fig:extra_samples}
\end{figure*}

We provide some additional examples of dataset samples in \Cref{fig:extra_samples}.


\end{document}